\documentclass[letterpaper, 10 pt, journal, twoside]{IEEEtran}
\usepackage[linesnumbered,ruled,vlined]{algorithm2e}
\usepackage{subfigure}
\usepackage{caption}
\usepackage{overpic}
\usepackage{stfloats}
\usepackage{graphics} 
\usepackage{graphbox} 
\usepackage{epsfig} 
\usepackage{mathptmx} 
\usepackage{times} 
\usepackage{amsmath} 
\usepackage{amssymb}  
\usepackage{amsbsy}
\usepackage{float}
\usepackage{mathrsfs} 
\DeclareMathAlphabet{\mathcal}{OMS}{cmsy}{m}{n} 
\usepackage{amsfonts}
\usepackage{bm}
\usepackage{cite}
\usepackage{hyperref} 
\usepackage{multirow}
\usepackage{siunitx} 
\usepackage{tikz}
\usepackage{algorithmic}
\usepackage[linewidth=1pt]{mdframed} 
\usepackage[export]{adjustbox} 
\usepackage{upgreek} 
\usepackage{booktabs}

\hypersetup{
	colorlinks=true,
	linkcolor=cyan, 
	filecolor=blue,      
	urlcolor=black,
	citecolor=black,
}

\begin{document}
\title{Hybrid Dynamics Modeling and Trajectory Planning for Cable-Trailer with Quadruped Robot System}
%
%
%

\author{Wentao Zhang$^{1}$,~\IEEEmembership{Graduate Student Member,~IEEE}, Shaohang Xu$^{1,2}$, Gewei Zuo$^{1}$, Bolin Li$^{1}$, Jingbo Wang$^{3}$ and Lijun Zhu$^{1}$,~\IEEEmembership{Member,~IEEE}%
\thanks{Manuscript received July, 6, 2025; Revised September, 30, 2025; Accepted October, 30, 2025.}
\thanks{This paper was recommended for publication by Editor Chung, Soon-Jo upon evaluation of the Associate Editor and Reviewers' comments.
This work was supported in part by the National Science and Technology Major Project under Grant 2022ZD0120003, the National Natural Science Foundation of China under Grant 62173155 and Grant 52188102, and in part by the Taihu Lake Innovation Fund for Future Technology, HUST. (Corresponding author: Lijun Zhu.)} 
\thanks{$^{1}$Wentao Zhang, Shaohang Xu, Gewei Zuo, Bolin Li, and Lijun Zhu are with the School of Artificial Intelligence and Automation, Huazhong University of Science and Technology, Wuhan 430074, China 
        {\tt\footnotesize wentaozhang@hust.edu.cn, shaohangxu@hust.edu.cn, gwzuo@hust.edu.cn, bolin\_li@hust.edu.cn, ljzhu@hust.edu.cn}}%
\thanks{$^{2}$Shaohang Xu is with the Department of Data Science, City University of Hong Kong, 999077, HKSAR
        {\tt\footnotesize shaohanxu2-c@my.cityu.edu.hk}}%
\thanks{$^{3}$Jingbo Wang is with the Embodied AI Center of Shanghai AI Lab, Shanghai, 200030, China
        {\tt\footnotesize wangjingbo1219@gmail.com}}%
\thanks{Digital Object Identifier (DOI): see top of this page.}
}
%
%

\markboth{IEEE Robotics and Automation Letters. Preprint Version. Accepted October, 2025}
{Zhang \MakeLowercase{\textit{et al.}}: Hybrid Dynamics Modeling and Trajectory Planning for Cable-Trailer with Quadruped Robot System} 

%



\maketitle
\begin{abstract}
Inspired by sled-pulling dogs in transportation, we present a cable-trailer integrated with a quadruped robot system. The motion planning of this system faces challenges due to the interactions between the cable's state transitions, the trailer's nonholonomic constraints, and the system's underactuation. To address these challenges, we first develop a hybrid dynamics model that captures the cable's taut and slack states. A search algorithm is then introduced to compute a suboptimal trajectory while incorporating mode transitions. Additionally, we propose a novel collision avoidance constraint based on geometric polygons to formulate the trajectory optimization problem for the hybrid system. The proposed method is implemented on a Unitree A1 quadruped robot with a customized cable-trailer and validated through experiments. The real system demonstrates both agile and safe motion with cable mode transitions. 
\end{abstract}


%
\IEEEpeerreviewmaketitle

\section{Introduction}
\subsection{Motivation}
\IEEEPARstart{Q}{uadruped} robots have demonstrated remarkable agility in navigating complex scenarios \cite{miki2022learning,bjelonic2022offline,jenelten2024dtc}, finding broad applications in fields such as search and rescue, factory inspection, and last-mile transportation. In transportation applications, a common approach involves directly fixing the load onto the robot's back. However, the quadruped robot's maximum load capacity is limited, and weight-bearing tasks are energy-intensive. This not only increases application costs but also restricts the robot's motion flexibility.

\begin{figure}[tbp] %
	\centering
	\subfigure[CT-QR System]{
		\includegraphics[width = 4.0 cm]{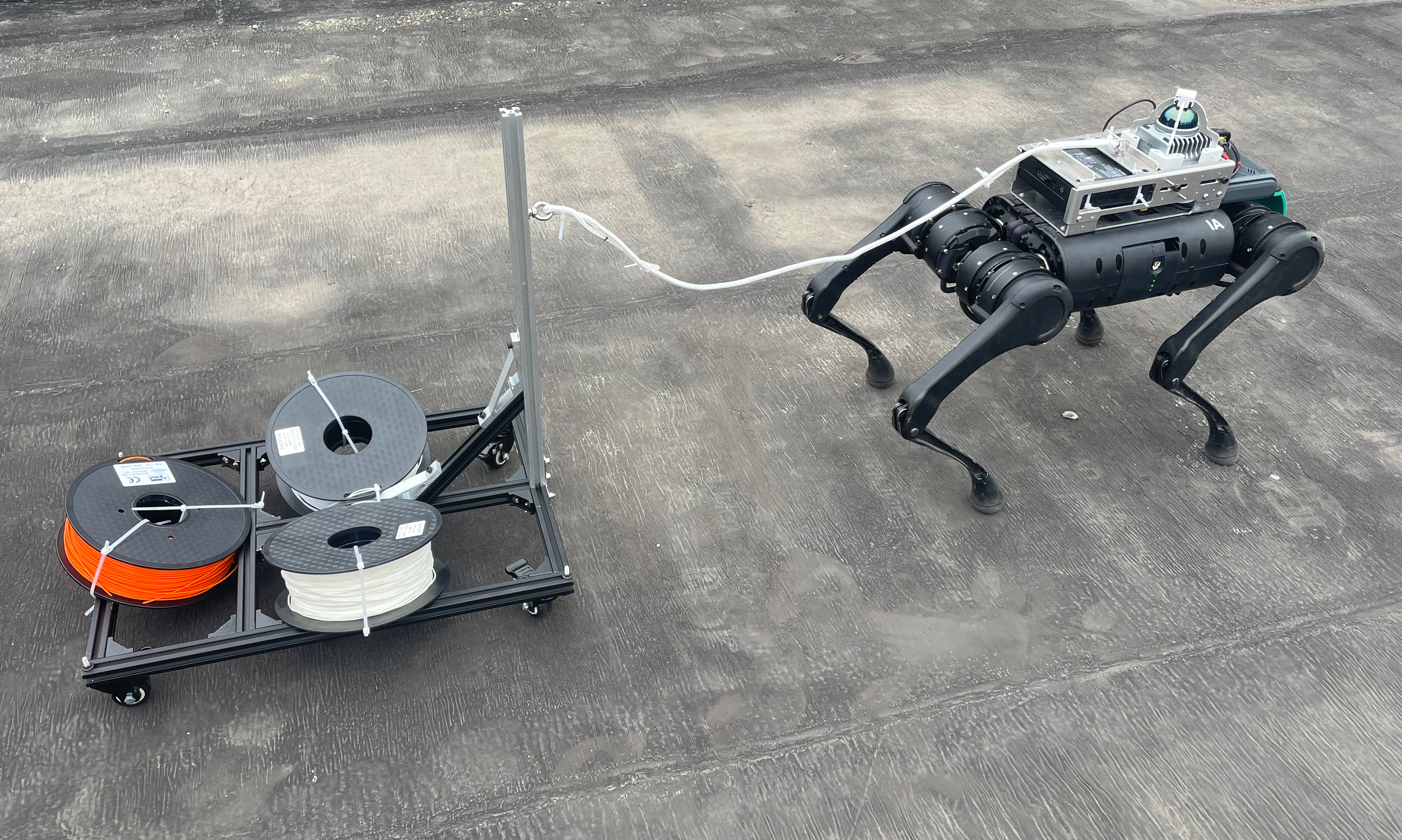}
		\label{fig:robotsled} %
	}\hspace{-0.2cm}
	\subfigure[System Configuration]{
		\includegraphics[width = 4.0 cm]{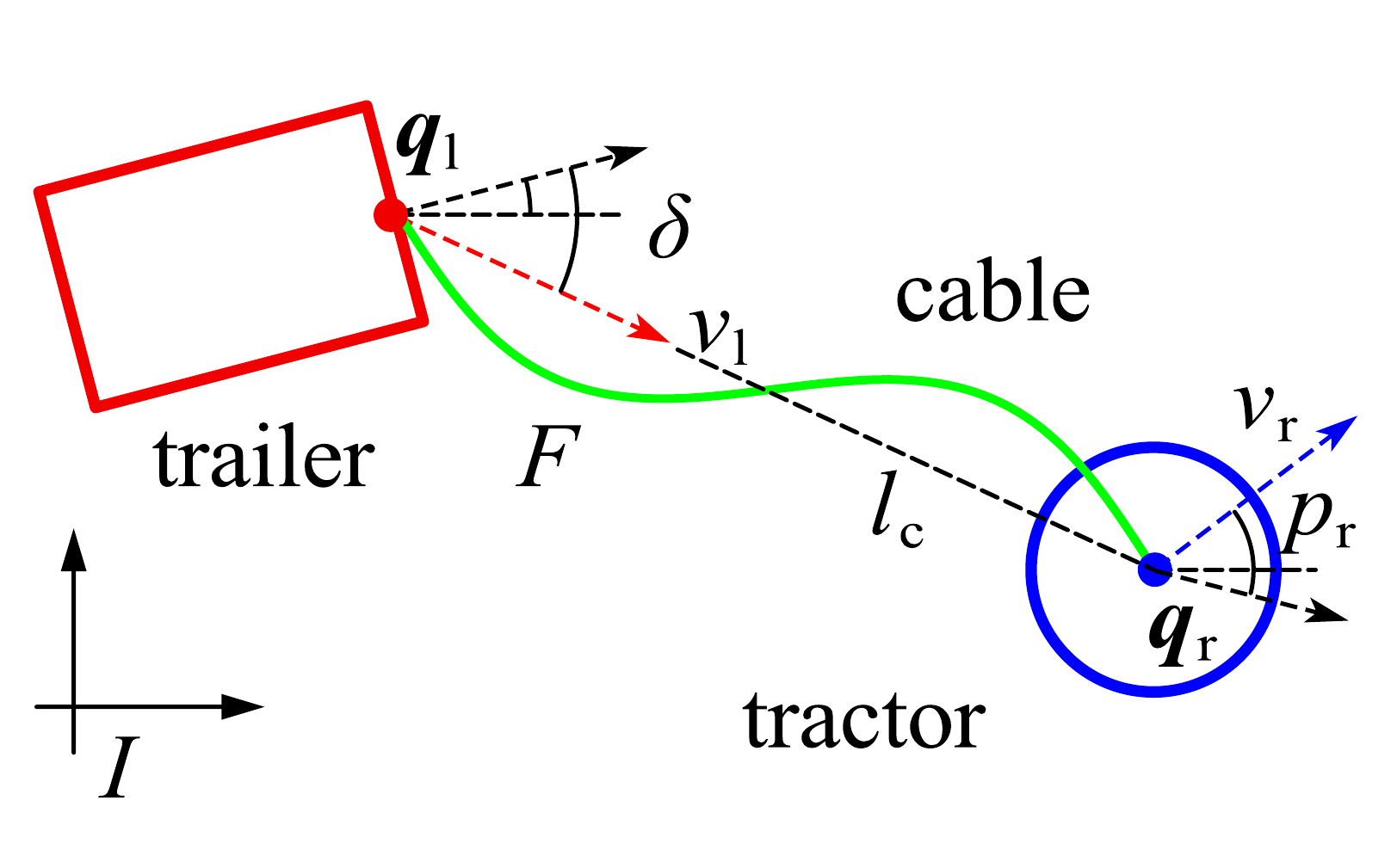}
		\label{fig:sysconfig} %
	}
	\subfigure[Motion Planning in Outdoor Environment]{
		\includegraphics[width = 8.2 cm]{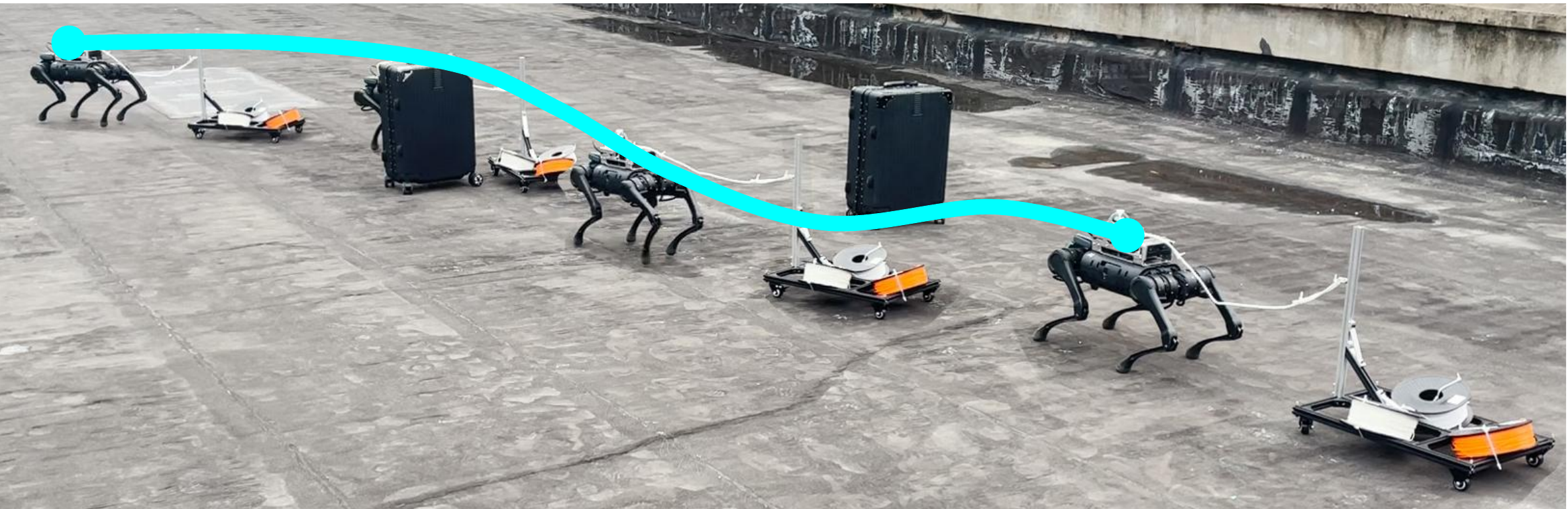}
		\label{fig:outdoortracking} %
	}
	\subfigure[Motion Planning in Narrow Corridor]{
		\includegraphics[width = 8.2 cm]{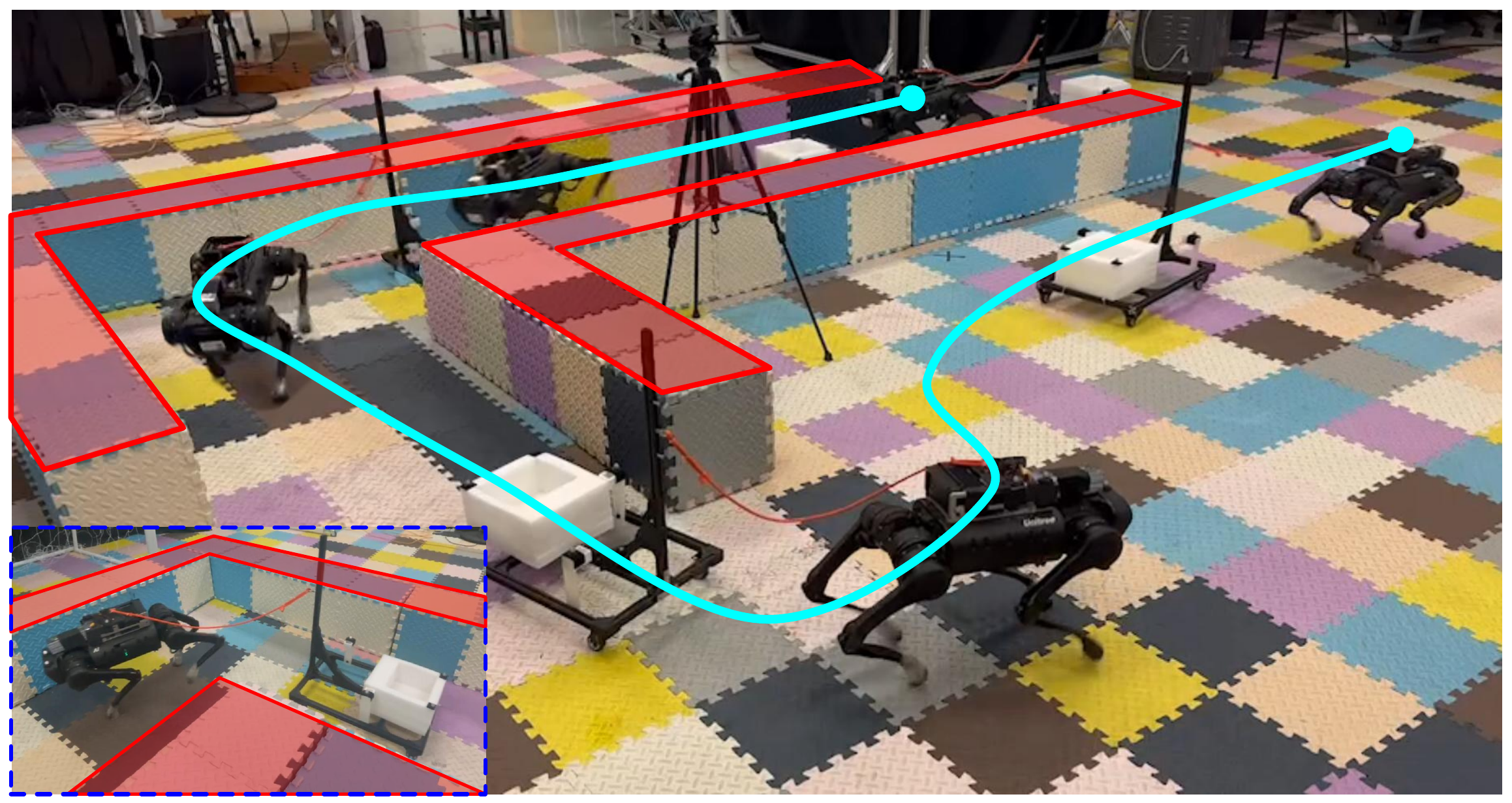}
		\label{fig:narrowexperiment} %
	}
	\caption{(a) illustrates the CT-QR system. (b) shows the configuration of the CT-QR system, where one end of the cable (green curve) is fixed at the tractor's center (blue circle), and the other end is fixed at the front of the wheeled platform (red rectangle). (c) presents motion planning experiment in outdoor environments. (d) presents real-world motion planning experiment in narrow corridor.}
	\label{fig:sledsystem}
\end{figure}

Inspired by sled-pulling dogs, we propose the cable-trailer with quadruped robot (CT-QR) system, shown in Fig. \ref{fig:robotsled}. The system consists of three parts: a quadruped robot(tractor), a soft cable, and a wheeled load platform(trailer). Compared to directly carrying loads, the CT-QR system is more energy-efficient due to the wheeled locomotion of trailer. Notably, the tractor is connected to the trailer by a flexible cable, and by leveraging the cable's tension and relaxation transition, the system can dynamically reshape and navigate cluttered environments.

However, motion planning for the CT-QR system is challenging. First, the nonholonomic constraints of the wheeled platform, combined with the system's underactuation, create complex constraints. Second, the cable's mode switching introduces hybrid system dynamics, complicating trajectory optimization and increasing computational complexity. 

In this paper, we develop a hybrid model for the CT-QR system and propose a planning framework that integrates hybrid search with trajectory optimization to leverage state transitions for safe and agile motion. We then provide a brief review of related work and outline our contributions.
	
\subsection{Related Work}
The car-like tractor with multiple trailers, serially connected by single joints and rigid links, have been discussed in \cite{ljungqvist2019path, li2019tractor, zhao2019modelling}. While these systems are underactuated and nonholonomic, they do not incorporate hybrid state transitions. The mobile robot with a cable-load system is explored in \cite{yu2022aggressive, foehn2017fast, zeng2020differential, yang2022collaborative, tang2015mixed, xiao2021robotic}. Specifically, \cite{yu2022aggressive} simplifies motion planning by maintaining taut cable tension, while \cite{foehn2017fast} models the cable-suspended payload as an additional link to a quadrotor, omitting hybrid modes. In \cite{zeng2020differential}, cable state is treated as a complementarity constraint within path planning. \cite{yang2022collaborative} tackles collaborative towing with cables by fixing the cable's hybrid mode in the real-time planning loop, aiming to optimize solutions by solving $2^n$ ($n$ being the number of robots) optimization problems. The approach in \cite{tang2015mixed} leverages a Mixed Integer Quadratic Program (MIQP) for hybrid trajectory generation, whereas \cite{xiao2021robotic} introduces a robot-guided human system, formulating a mixed-integer problem but neglecting trailer dynamics.

It is important to note that the CT-QR system differs from the above work. The wheeled platform (trailer) is nonholonomic and underactuated, exhibiting complex dynamic characteristics. Our motion planning objective is to generate agile and safe high-speed trajectories that guide the trailer to its target pose. Moreover, the cable's mode transitions are explicitly considered to handle complex environments.

The two-stage planning framework provides a systematic approach to addressing complex motion-planning problems. For example, \cite{li2019tractor} demonstrates flexible motion planning for tractor-trailer vehicles in narrow environments, \cite{zhang2022autonomous} proposes a hierarchical method for autonomous navigation of terrestrial-aerial bimodal vehicles, and \cite{sleiman2023versatile} presents a versatile multi-contact planning and control framework for quadruped robots with manipulators. We adopt this framework for the CT-QR system to manage complex hybrid mode transitions and generate agile and safe trajectories efficiently.

Whole-Body collision avoidance is modeled as hard constraints in \cite{zhang2020optimization,li2015unified,yang2021whole} and soft constraints in \cite{wang2023linear,geng2023robo,zhang2023continuous}. \cite{zhang2020optimization} presents a method that reformulates nondifferentiable collision avoidance constraints into smooth, differentiable forms using the strong duality of convex optimization. \cite{li2015unified} introduces a rectangle-shaped collision avoidance approach based on shape area calculations. \cite{yang2021whole} constructs flight corridors and robot convex polyhedrons as inequality constraints to ensure collision avoidance. \cite{wang2023linear} models both obstacles and robots as convex polyhedrons, employing half-space evaluation for collision detection. \cite{geng2023robo} proposes the Robo-centric Euclidean Signed Distance Field (RC-ESDF) with a penalty function for effective collision avoidance. \cite{zhang2023continuous} models the robot as an implicit function, using swept volume to evaluate continuous collisions, even for nonconvex robots with complex surfaces. 

Soft constraints introduce potential collision risks, while hard constraints are either conservative or computationally expensive. To expand the application of CT-QR system in complex environments, we propose a novel  formulation of geometry collision avoidance constraint that fully accounts for the variable shape of objects, offering both aggressiveness and computational efficiency.

\subsection{Contributions}
In this paper, we present a novel CT-QR system and design a hierarchical planning framework to guide the trailer to target poses with agile and safe maneuvers in complex environments. The specific contributions are as follows:
\begin{itemize}
	\item A hybrid dynamics model is introduced to capture the dynamic characteristics of the CT-QR system, including nonholonomic constraints, underactuation, and hybrid cable-state transitions. 
	\item A hierarchical planning framework is proposed for hybrid system trajectory planning. The front-end hybrid search generates sub-optimal trajectories with mode transitions, which serves as the initial guess for the back-end trajectory optimization. The back-end module then refines these trajectories to be energy-efficient, dynamics feasible, and safe. In particular, we introduce an efficient inequality constraint to address the variable shape of the CT-QR system. 
	\item The proposed CT-QR system, along with the planning method, is validated through experiments, demonstrating the effectiveness and efficiency of our approach. 
\end{itemize}

\subsection{Structure}
The remainder of this article is organized as follows: Section \ref{sec:hybridmodel} introduces the dynamics model of the CT-QR system. In Section \ref{sec:trajectory}, we detail the trajectory generation and optimization approach. Section \ref{sec:experiments} presents an analysis and evaluation of our method, supported by numerical simulations and real-world experiments. Finally, Section \ref{sec:conclusion} provides the concluding remarks.

\section{Hybrid Dynamics Model of CT-QR System}\label{sec:hybridmodel}
\subsection{System Configuration} 
We consider the motion of the CT-QR system in the plane, using $SE\left( 2 \right) $ to represent the states of the tractor $\boldsymbol{q}_{\mathrm{r},k}=\left[ x_{\mathrm{r},k},y_{\mathrm{r},k},\theta_{\mathrm{r},k} \right]^{\top}$ and trailer $\boldsymbol{q}_{\mathrm{l},k}=\left[ x_{\mathrm{l},k},y_{\mathrm{l},k},\theta_{\mathrm{l},k} \right]^{\top}$. Let $I$ denote the world inertial coordinate system. We define the direction vector in $I$ as $\boldsymbol{e}\left( \theta \right)=\left[ \cos \left( \theta \right) ,\sin \left( \theta \right) \right]^{\top}$. The configuration of the CT-QR system is shown in Fig. \ref{fig:sysconfig}. The system's state and input are defined as 
\begin{equation*}
	\begin{split}
	\boldsymbol{x}_{k} &= \left[ \boldsymbol{q}_{\mathrm{l},k},\boldsymbol{q}_{\mathrm{r},k},\dot{\boldsymbol{q}}_{\mathrm{l},k},\dot{\boldsymbol{q}}_{\mathrm{r},k},\delta_{k},v_{\mathrm{l},k},F_{k},\eta_{k} \right] ^{\top}\in \mathbb{R} ^{16}
	\\
	\boldsymbol{u}_{k}&=\ddot{\boldsymbol{q}}_{\mathrm{r},k}=\left[ a_{\mathrm{r},k}^{\mathrm{x}},a_{\mathrm{r},k}^{\mathrm{y}},g_{\mathrm{r},k} \right] ^{\top}\in \mathbb{R} ^3,
	\end{split}
\end{equation*}
where $\delta_{k}$ represents trailer's front wheel steering angle, $v_{\mathrm{l},k}$ denotes trailer's velocity, and $F_{k}$ represents cable force. $\eta_{k}$ defines the hybrid state of the cable, with $\eta_{k}=1$ for taut and $\eta_{k}=0$ for slack. $a_{\mathrm{r},k}^{\mathrm{x}}$ and $a_{\mathrm{r},k}^{\mathrm{y}}$ are tractor's linear accelerations along the $x$-axis and $y$-axis respectively, and $g_{\mathrm{r},k}$ represents tractor's angular acceleration. The physical parameters of the CT-QR system, shown in Fig. \ref{fig:robotsled}, are listed in Tab. \ref{tab:sysconfig}.

\begin{table}[htbp]
	\renewcommand{\arraystretch}{1.3}
	\centering
	\caption{CT-QR SYSTEM CONFIGURATION}
	\label{tab:sysconfig}
	\begin{tabular}{clc}
	\hline
	\textbf{Symbol}& \multicolumn{1}{c}{\textbf{Description}} & \textbf{Value} \\  
	\hline
	$\varPhi$ & Maximum steering angle of trailer & \SI{90}{\degree} \\ 
	$L_{\mathrm{l}}$ & Distance between rear wheel and front wheel & \SI{0.5}{m} \\ 
	$L_{\mathrm{c}}^{\mathrm{ub}}$ & Maximum cable length & \SI{0.8}{m} \\
	$L_{\mathrm{c}}^{\mathrm{lb}}$ & Minimum cable length & \SI{0.2}{m} \\
	$L_{\mathrm{s}}$ & Safe distance between trailer and tractor & \SI{0.55}{m} \\
	$\mu$ & Trailer wheel friction coefficient & 0.03 \\
	\hline
	\end{tabular}
\end{table}
\subsubsection{Tractor Configuration} 
We model the omnidirectional motion of the tractor using a double-integral mass-point model, which can also accommodate various types of omnidirectional mobile robots. The tractor's input is denoted as  $\boldsymbol{u}_{\mathrm{r},k}=\boldsymbol{u}_{k}$, and its state is $\boldsymbol{x}_{\mathrm{r},k}=\left[ \boldsymbol{q}_{\mathrm{r},k}^{\top},\dot{\boldsymbol{q}}_{\mathrm{r},k}^{\top} \right] ^{\top}\in \mathbb{R} ^6$. The discretized dynamics model of tractor is described as 
\begin{equation}
	\begin{split}
		\left[ \begin{array}{c}
			\boldsymbol{q}_{\mathrm{r},k+1}\\
			\dot{\boldsymbol{q}}_{\mathrm{r},k+1}\\
		\end{array} \right] &=\left[ \begin{array}{c}
			\dot{\boldsymbol{q}}_{\mathrm{r},k}\\
			\ddot{\boldsymbol{q}}_{\mathrm{r},k}\\
		\end{array} \right] \cdot \mathrm{d}t+\left[ \begin{array}{c}
			\boldsymbol{q}_{\mathrm{r},k}\\
			\dot{\boldsymbol{q}}_{\mathrm{r},k}\\
		\end{array} \right],
	\end{split}
	\label{equ:tractormotion}
\end{equation}
where $\mathrm{d}t$ represents the discrete time step, and $k$ denotes the discrete-time index.

\subsubsection{Trailer Configuration}
\begin{figure}[htbp] 
	\centering
	\includegraphics[width = 7.2 cm]{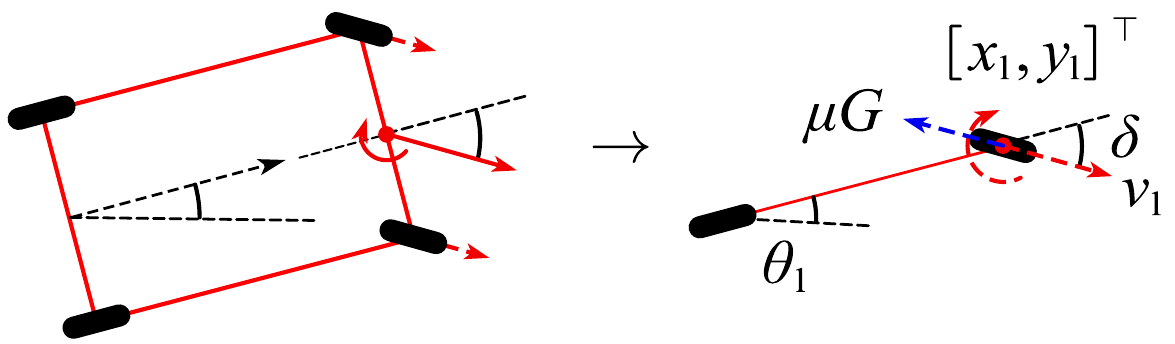}
	\caption{Configuration of trailer, which has two rear wheels with fixed directions and two front wheels for steering.}
	\label{fig:trailer}
\end{figure}

The trailer is driven by the force from the cable and has zero lateral velocity (sideslip is not considered), as shown in Fig. \ref{fig:trailer}. This leads to nonholonomic constraints and underactuation. Inspired by the bicycle model in~\cite{ge2021numerically}, we define the discretized trailer dynamics model as
\begin{equation}
	\begin{split}
		\left[ \hspace{-0.10cm} \begin{array}{c}
			x_{\mathrm{l},k+1}\\
			y_{\mathrm{l},k+1}\\
			\theta _{\mathrm{l},k+1}\\
			v_{\mathrm{l},k+1}\\
		\end{array} \hspace{-0.10cm}\right] \hspace{-0.05cm} &=\hspace{-0.05cm}\left[\hspace{-0.10cm} \begin{array}{c}
			v_{\mathrm{l},k}\cdot \cos \left( \theta _{\mathrm{l},k}+\delta _{k} \right)\\
			v_{\mathrm{l},k}\cdot \sin \left( \theta _{\mathrm{l},k}+\delta _{k} \right)\\
			v_{\mathrm{l},k}\cdot \sin \left( \delta _{k} \right) /L_{\mathrm{l}}\\
			{{F_{k}} / {m_{\mathrm{l}}}}-\mu G\cdot \mathrm{sgn} \left( v_{\mathrm{l},k} \right)\\
		\end{array} \hspace{-0.10cm}\right] \cdot \mathrm{d}t+\left[\hspace{-0.10cm} \begin{array}{c}
			x_{\mathrm{l},k}\\
			y_{\mathrm{l},k}\\
			\theta _{\mathrm{l},k}\\
			v_{\mathrm{l},k}\\
		\end{array} \hspace{-0.10cm}\right],
	\end{split}
	\label{equ:trailermotion}
\end{equation}
where the trailer state is $\boldsymbol{x}_{\mathrm{l},k}=\left[ x_{\mathrm{l},k},y_{\mathrm{l},k},\theta _{\mathrm{l},k},v_{\mathrm{l},k} \right] ^{\top}\in \mathbb{R} ^4$, and the trailer input is $\boldsymbol{u}_{\mathrm{l},k}=\left[ \delta_k,F_k \right] ^{\top}\in \mathbb{R} ^2$. $G$ is the gravitational acceleration, and $m_{\mathrm{l}}$ is trailer's mass. We assume that the ground friction is constant and acts in the direction opposite to the trailer's velocity.

\subsubsection{Cable Configuration}
The cable is assumed to be non-stretchable and massless, with related states described as
\begin{equation*}
	\begin{split}
		l_{\mathrm{c},k} &=\left\| \left[ x_{\mathrm{r},k}-x_{\mathrm{l},k},y_{\mathrm{r},k}-y_{\mathrm{l},k} \right] \right\|
		\\
		\theta _{\mathrm{c},k}&=\mathrm{arctan} \left( \frac{y_{\mathrm{r},k}-y_{\mathrm{l},k}}{x_{\mathrm{r},k}-x_{\mathrm{l},k}} \right)
		\\
		\omega _{\mathrm{c},k}&=v_{\mathrm{r},k}\cdot \sin \left( p_{\mathrm{r},k}-\theta _{\mathrm{c},k} \right) /l_{\mathrm{c},k},
	\end{split}
	\label{equ:cablemotion}
\end{equation*}
where $l_{\mathrm{c},k}$ represents cable length, $\theta _{\mathrm{c},k}$ denotes cable direction angle, and $\omega _{\mathrm{c},k}$ is cable angular velocity. $v_{\mathrm{r},k}=\left\| \left[ \dot{x}_{\mathrm{r},k},\dot{y}_{\mathrm{r},k} \right] \right\|$ is tractor linear velocity and $p_{\mathrm{r},k}=\mathrm{arctan} \left( \dot{y}_{\mathrm{r},k}/\dot{x}_{\mathrm{r},k} \right)$ denotes its direction angle. These state variables are used to facilitate the formulation and calculation of the CT-QR system dynamics.

\subsection{Hybrid Dynamics Model}
\begin{figure}[htbp]
	\center
	\subfigure[Slack Model]{
		\includegraphics[height = 2.6 cm]{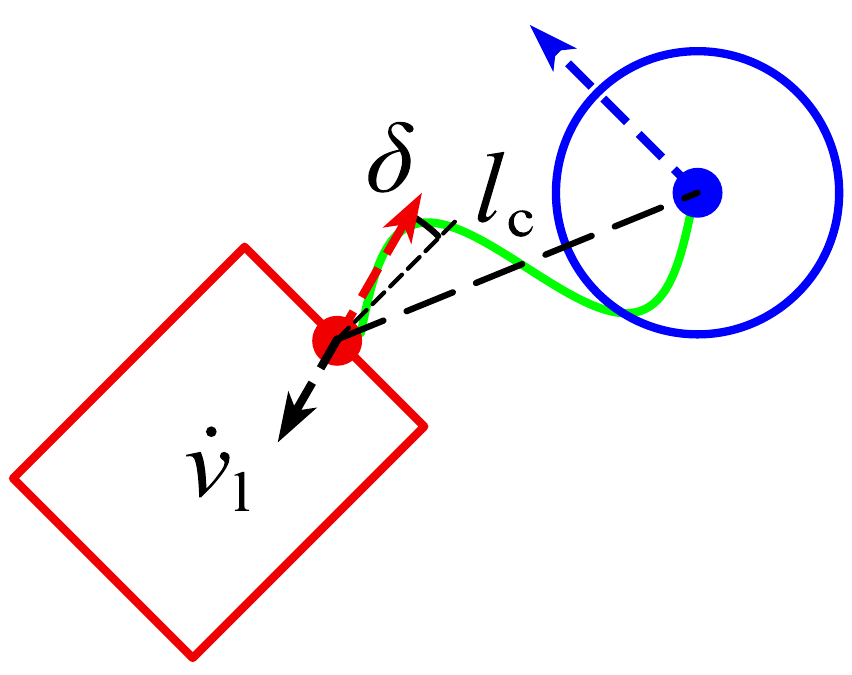}
		\label{fig:slackmodel} %
	}
	\subfigure[Taut Mode]{
		\includegraphics[height = 2.6 cm]{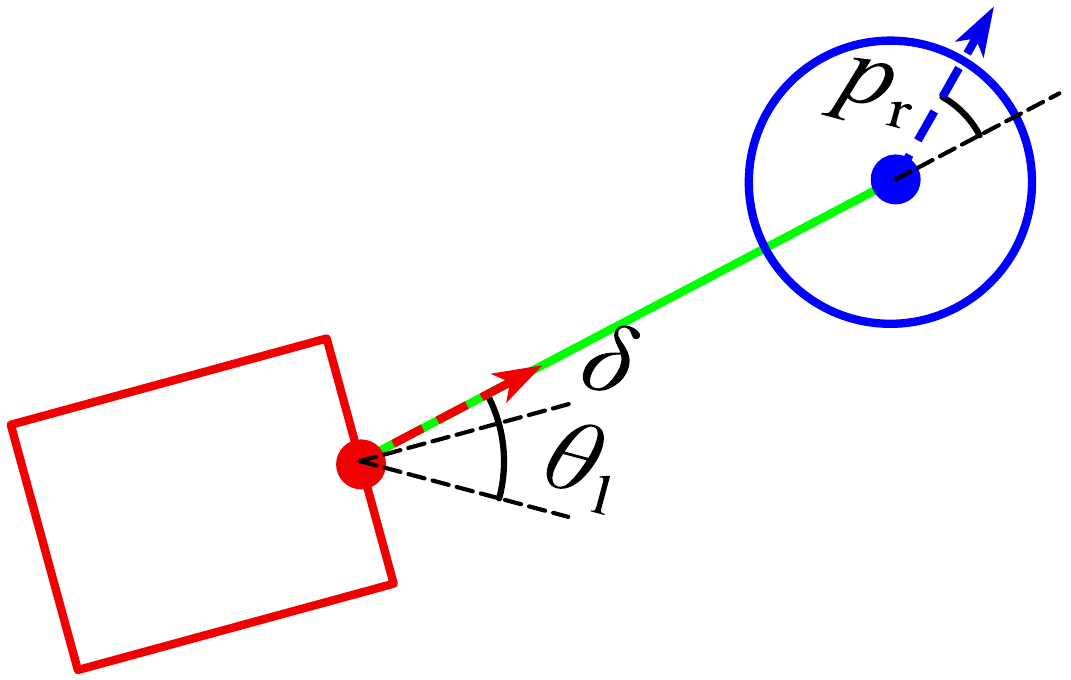}
		\label{fig:tautmodel} %
	}
	\caption{CT-QR system hybrid motion modes.}
	\label{fig:hybridmodel}
\end{figure}

We formulate the taut and slack motion constraints of the soft cable separately and integrate them with the tractor and trailer model to describe the hybrid dynamics of the CT-QT system, as shown in Fig. \ref{fig:hybridmodel}.
\subsubsection{Slack Model}
When the cable is slack, the trailer enters a free deceleration state with $F_{k}=0$, as defined in \eqref{equ:vlecon}, and the steering angle is fixed with $\dot{\delta_{k}} =0$, as defined in \eqref{equ:flecon}, due to the tendency to maintain the momentum of the previous motion. Furthermore, the cable length is within the range $L_{\mathrm{c}}^{\mathrm{lb}}\leqslant l_{\mathrm{c},k}\leqslant L_{\mathrm{c}}^{\mathrm{ub}}$, as referenced in \eqref{equ:lcieconUB} and \eqref{equ:lcieconLB}, and the motion of tractor and trailer do not interfere with each other. The details of the slack constraints are shown below:  
\begin{subequations}
	\begin{align}
		\eta_{k} &=0 
		\\
		v_{\mathrm{l},k}-\mu G\cdot \mathrm{sgn} \left( v_{\mathrm{l},k} \right) -v_{\mathrm{l},k+1} &=0 \label{equ:vlecon}
		\\
		\delta _{k+1}-\delta _{k} &=0 \label{equ:flecon}
		\\
		l_{\mathrm{c},k}^{2}-{L_{\mathrm{c}}^{\mathrm{ub}}}^2 &\leqslant 0 \label{equ:lcieconUB} 
		\\
		{L_{\mathrm{c}}^{\mathrm{lb}}}^2-l_{\mathrm{c},k}^{2} &\leqslant 0
		\label{equ:lcieconLB}
	\end{align}
	\label{equ:slackconstraint}
\end{subequations}
The slack dynamics model, consisting of \eqref{equ:tractormotion}, \eqref{equ:trailermotion} and \eqref{equ:slackconstraint}, is denoted as $\boldsymbol{x}_{k+1}=f_{\mathrm{s}}\left( \boldsymbol{x}_{k},\boldsymbol{u}_{k},\mathrm{d}t \right)$. Notably, the mode constraints associate the tractor and trailer dynamics.

\subsubsection{Taut Model}
In taut mode, the tractor pulls the trailer through the cable, which reaches its maximum length $l_{\mathrm{c},k}=L_{\mathrm{c}}^{\mathrm{ub}}$ \eqref{equ:lcecon}, and the steering angle aligns with the cable direction, $\delta_k=\theta_{\mathrm{c},k}-\theta_{\mathrm{l},k}$, as shown in \eqref{equ:flqlecon} and \eqref{equ:flqliecon}. Additionally, there is a tendency for tractor and trailer to move away from each other, with the condition ${{\pi}/{2}}\geqslant \left| p_{\mathrm{r},k}-\theta _{\mathrm{c},k} \right|$, since the direction of linear velocity of tractor forms an acute angle with the cable. The details of the taut constraints are shown below:
\begin{subequations}
	\begin{align}
		\eta_{k} &=1
		\\
		L_{\mathrm{c}}^{\mathrm{ub}2}-l_{\mathrm{c},\mathrm{k}}^{2} &=0 \label{equ:lcecon}
		\\
		\boldsymbol{e}\left( \theta _{\mathrm{l},k}+\delta _k \right) \times \left[ x_{\mathrm{r},k}-x_{\mathrm{l},k},y_{\mathrm{r},k}-y_{\mathrm{l},k} \right] ^{\top} &=0 \label{equ:flqlecon}
		\\
		-\boldsymbol{e}\left( \theta _{\mathrm{l},k}+\delta _k \right) ^{\top}\cdot \left[ x_{\mathrm{r},k}-x_{\mathrm{l},k},y_{\mathrm{r},k}-y_{\mathrm{l},k} \right] ^{\top} &\leqslant0 \label{equ:flqliecon}
		\\
		-\boldsymbol{e}\left( \theta _{\mathrm{l},k}+\delta _k \right) ^{\top}\cdot \left[ \dot{x}_{\mathrm{r},k},\dot{y}_{\mathrm{r},k} \right] ^{\top} &\leqslant 0. \label{equ:vriecon}
	\end{align} 
	\label{equ:tautconstraint}
\end{subequations}
The taut dynamics model is denoted as $\boldsymbol{x}_{k+1}=f_{\mathrm{t}}\left( \boldsymbol{x}_{k},\boldsymbol{u}_{k},\mathrm{d}t \right)$, which includes \eqref{equ:tractormotion}, \eqref{equ:trailermotion}, and \eqref{equ:tautconstraint}.

\section{Trajectory Generation and Optimization}\label{sec:trajectory}
The proposed framework leverages the cable's taut/slack state transitions to guide the trailer to the target pose. The front-end module generates feasible, sub-optimal trajectories with mode transitions, while the back-end module refines these trajectories to local optimality, incorporating collision avoidance with variable geometry shapes. 

\subsection{Hybrid Search}\label{sec:search}
The mode transition of the soft cable depends on the control input and the current state, often requiring more than one time step to transition from slack mode to taut mode. Therefore, traditional search methods, which rely on inverse calculations from state space or workspace to determine the control input, are questioned in the CT-QR system. Inspired by the hybrid A* search \cite{dolgov2010path}, we incorporate state transitions into forward dynamics iterations and propose the hybrid search module. The main components of this module are outlined in Alg.~\ref{alg:search}, with further details can be found in \cite{dolgov2010path}. Notably, our method generates long-term slack and taut mode trajectory segments with natural and smooth transitions.

$\mathcal{N}$ and $\mathcal{C}$ represent the open and closed sets, respectively. \textbf{Feasible}($\cdot$) checks the safety of the node states. $f_\mathrm{c}$ denotes the final cost of a node, while $g_\mathrm{c}$ and $h_\mathrm{c}$ refer to the edge cost and heuristic cost, calculated by \textbf{EdgeCost}($\cdot$) and \textbf{Heuristic}($\cdot$), respectively.

\IncMargin{1em}
\begin{algorithm}[htbp] 
	\SetKwData{Primitives}{$\mathcal{U}$}
	\SetKwData{Uinput}{$\boldsymbol{u}_{\mathrm{in}}$}
	\SetKwData{Nodet}{$n_{\mathrm{t}}$}
	\SetKwData{Nodei}{$n_{\mathrm{i}}$}
	\SetKwData{Nodec}{$n_{\mathrm{c}}$}
	\SetKwData{State}{$\boldsymbol{x}$}
	\SetKwData{StateIn}{$\boldsymbol{x}_{\mathrm{in}}$}
	\SetKwData{StateOut}{$\boldsymbol{x}_{\mathrm{out}}$}
	\SetKwData{gnCost}{$g_{\mathrm{c}}$}
	\SetKwData{hnCost}{$h_{\mathrm{c}}$}
	\SetKwData{fnCost}{$f_{\mathrm{c}}$}
	\SetKwData{Parent}{$parent$}
	\SetKwData{dT}{$T$}
	\SetKwData{dt}{${\mathrm{d}}t$}
	\SetKwData{Open}{$\mathcal{N}$}
	\SetKwData{Closed}{$\mathcal{C}$}

	\SetKwFunction{Initialize}{Initialize()}
	\SetKwFunction{Empty}{{\bf empty}}
	\SetKwFunction{Pop}{{\bf pop}}
	\SetKwFunction{Push}{{\bf push}}
	\SetKwFunction{Contain}{{\bf contain}}
	\SetKwFunction{Insert}{{\bf insert}}
	\SetKwFunction{Expand}{{\bf Expand}}
	\SetKwFunction{Update}{{\bf Update}}
	\SetKwFunction{FTaut}{$f_{\mathrm{t}}$}
	\SetKwFunction{FSlack}{$f_{\mathrm{s}}$}
	\SetKwFunction{EdgeCost}{{\bf EdgeCost}}
	\SetKwFunction{Heuristic}{{\bf Heuristic}}
	\SetKwFunction{Check}{{\bf Feasible}}
	\SetKwFunction{Reach}{{\bf ReachGoal}}
	\SetKwFunction{GetPath}{{\bf GetPath}}
	\SetKwFunction{Continue}{{\bf continue}}

	\While{$\neg \;$ \Open.\Empty{}} {
	\Nodec $\leftarrow$ \Open.\Pop{}, \Closed.\Push{}\\
	\If{\Reach{\Nodec}} {
		\Return{\GetPath{\Nodec, \Closed}}\\
	}
	\For{\Uinput $\leftarrow$ \Primitives} {
			\StateIn $\leftarrow$ \Nodec.\State\\
			\For{$i\leftarrow 1$ \KwTo $T_{\mathrm{s}}/\mathrm{d}t$} {
				\StateOut $\leftarrow$ \FSlack(\StateIn, \Uinput, \dt)\\
				\If{\StateOut.$l_{\mathrm{c}} > {L_{\mathrm{c}}^{\mathrm{ub}}}$} {
					\StateOut $\leftarrow$ \FTaut{\StateIn, \Uinput, \dt}\\
				}
				\Nodet.\Push{\StateOut}; \\
				\StateIn $\leftarrow$ \StateOut;
			}
			\If{$\neg \;$ \Closed.\Contain{\Nodet} $\wedge$ \Check{\Nodet}} {
				\Nodet.\gnCost $\leftarrow$ \Nodec.\gnCost + \EdgeCost{\Nodet};\\
				\Nodet.\fnCost $\leftarrow$ \Nodet.\gnCost $\,$+ \Heuristic{\Nodet};\\
				\Open.\Update{\Nodet}\\
			}
		}
	}
	\caption{Hybrid Dynamics Trajectory Search}
	\label{alg:search}
\end{algorithm}
\DecMargin{1em}

\subsubsection{Node Expanding}
The discretized control input set is defined as
\begin{equation*}
	\mathcal{U} =\left\{ \left[ a\cos \left( \theta \right) ,a\sin \left( \theta \right) \right] ^{\top}|\forall a\in \mathcal{U} _{\mathrm{a}},\theta \in \mathcal{U} _{\uptheta} \right\},
\end{equation*} 
where $\mathcal{U} _{\mathrm{a}}=\left\{ 0,D_{\mathrm{a}},...,a_{\max} \right\}$, with the maximum linear acceleration $a_{\max}$ and acceleration resolution $D_{\mathrm{a}}$. Additionally, $\mathcal{U} _{\uptheta}=\left\{ D_{\uptheta},2D_{\uptheta},...,2\pi \right\}$ and $D_{\uptheta}$ is direction resolution. The direction control of the tractor is neglected during the search process due to its omnidirectional motion capability, but it will be considered during the trajectory optimization process.

The system state sequences are computed iteratively using the dynamics functions $f_{\mathrm{t}}(\cdot)$ and $f_{\mathrm{s}}(\cdot)$, based on the given expansion period $T_{\mathrm{s}}$ and the current state $\boldsymbol{x}_{\mathrm{in}}$, the next state $\boldsymbol{x}_{\mathrm{out}}$ can be computed within two forward iterations, as shown in lines 7 to 12 of Alg. \ref{alg:search}.  Notably, the nonholonomic constraints, underactuation, and cable state transition are naturally handled in this process. The state grid for the hybrid A* algorithm is determined by tractor's workspace $[x_{\mathrm{r},k},y_{\mathrm{r},k}]$, trailer's workspace $[x_{\mathrm{l},k},y_{\mathrm{l},k},\theta_{\mathrm{l},k}]$, and cable state $\eta_{k}$, with distance resolution $D_{\mathrm{d}}$ and rotation resolution $D_{\mathrm{r}}$, as:
\begin{equation*}
	\begin{aligned}
		\left[ \frac{x_{\mathrm{r},k}}{D_{\mathrm{d}}},\, \frac{y_{\mathrm{r},k}}{D_{\mathrm{d}}},\, \frac{x_{\mathrm{l},k}}{D_{\mathrm{d}}},\, \frac{y_{\mathrm{l},k}}{D_{\mathrm{d}}},\, \frac{\theta _{\mathrm{l},k}}{D_{\mathrm{r}}},\, \eta_{k} \right] ^{\top}\in \mathbb{Z} ^6.
	\end{aligned}
\end{equation*}
\subsubsection{Edge Cost}
We aim to find a trajectory that is optimal in terms of both time and motion distance. The edge cost $g_{\mathrm{c}}$ is defined as:
\begin{equation*}
	g_{\mathrm{c}} =\lambda _{\mathrm{l}}l_{\mathrm{l}}+\lambda _{\mathrm{r}}l_{\mathrm{r}}+\lambda _{\mathrm{t}}T,
\end{equation*}
where $\lambda _{\mathrm{l}}$, $\lambda _{\mathrm{r}}$ and $\lambda _{\mathrm{t}}$ are the weights for trailer path length $l_{\mathrm{l}}$, tractor path length $l_{\mathrm{r}}$, and trajectory time duration $T$, respectively. The motion path length contains of horizontal displacement and yaw angle change, weighted by $\lambda _{\mathrm{a}}$ as
\begin{equation*}
	l=\sum_{k=1}^{T/\mathrm{d}t}{\left( \begin{array}{c}
		\left( 1-\lambda _{\mathrm{a}} \right) \cdot \left| \theta _{k}-\theta _{k-1} \right|\;\; +\;\;\\
		\lambda _{\mathrm{a}}\cdot \left\| \left[ x_{k}-x_{k-1},y_{k}-y_{k-1} \right] \right\|\\
	\end{array} \right)}.
\end{equation*}

\subsubsection{Heuristic Cost}
The original hybrid A* algorithm is guided by two heuristic terms: the \textit{holonomic-with-obstacles} cost $h_{\mathrm{a}}$ and the \textit{nonholonomic-without-obstacles} cost $h_{\mathrm{d}}$. In this paper, we retain the former to account for obstacle geometry and prevent convergence into local dead zones. The original $h_{\mathrm{d}}$ is the length of ReedsShepp curve \cite{reeds1990optimal}, which ignores obstacles but considers the nonholonomic constraints to prevent convergence towards unreachable directions. However, in the CT-QR system, the tractor can only drive the trailer when the cable is taut. Thus, we focus on the forward motion of the system, modeled by Dubins curve \cite{shkel2001classification} with the minimum turning radius $r_{\min}=\left\| L_{\mathrm{l}}+L_{\mathrm{c}}^{\mathrm{ub}}\cdot \cos \left( \varPhi \right) ,L_{\mathrm{c}}^{\mathrm{ub}}\cdot \sin \left( \varPhi \right) \right\| $. Finally, the heuristic cost is defined as $h_{\mathrm{c}}=\max \left( h_{\mathrm{a}},h_{\mathrm{d}} \right)$.

\subsubsection{Stopping Criterion}
The search process continues until all nodes in the open set are expanded or the goal is reached. To speed up the search process, we interpolate Dubins curve by fixing $\delta_{k} = 0$ and $\eta_{k} = 1$, to obtain taut mode state sequences, which are then checked for feasibility. This approach is implemented in \textbf{ReachGoal}($\cdot$), significantly enhancing the efficiency of the search process.

\subsection{Trajectory Optimization}
Based on the search trajectory $\varGamma$, the CT-QR system can be guided to the goal without collision. However, $\varGamma$ is a suboptimal trajectory due to sampling resolution and the lack of smoothness consideration. To improve the trajectory further, we formulate nonlinear trajectory optimization as \eqref{equ:NTO} to obtain energy-efficient, smooth, safe, and dynamic feasible trajectories. It is important to note that the original trajectory optimization problem for hybrid systems is nonlinear, non-convex, and may include discontinuities. $\varGamma$ serves as a high-quality initial guess for solving this problem.
\begin{subequations}
	\begin{align}
		\min_{\boldsymbol{x},\boldsymbol{u}} \,\,L(\boldsymbol{x},\boldsymbol{u},t) &=L_{\mathrm{R}}+L_{\mathrm{O}} \label{equ:objective}
		\\
		\mathrm{s}.\mathrm{t}.\quad \boldsymbol{x}_0&=\boldsymbol{x}_{\mathrm{start}}  \nonumber
		\\
		\boldsymbol{x}_{\mathrm{N}}&=\boldsymbol{x}_{\mathrm{goal}}  \nonumber
		\\
		\mathbf{if}\;\,\eta _k&=0\;\;\mathbf{then}  \nonumber
		\\
		\boldsymbol{x}_{k+1}&=f_{\mathrm{s}}\left( \boldsymbol{x}_k,\boldsymbol{u}_k,\mathrm{d}t \right) \nonumber
		\\
		\mathbf{else}\;\;&\mathbf{if}\;\;\eta _k=1  \nonumber
		\\
		\boldsymbol{x}_{k+1}&=f_{\mathrm{t}}\left( \boldsymbol{x}_k,\boldsymbol{u}_k,\mathrm{d}t \right) \label{equ:hybridstate} 
		\\
		h_{\mathrm{b}} \left( \boldsymbol{x}_k,\boldsymbol{u}_{k} \right) &\leqslant 0 \label{equ:inequalbound}
		\\
		h_{\mathrm{o}} \left( \boldsymbol{x}_k,\mathcal{O} \right) &\geqslant 0 \label{equ:obstacle}
	\end{align}
	\label{equ:NTO}
\end{subequations}
\subsubsection{Cost Function}
The input and state cost $L_{\mathrm{R}}$ penalizes control input and state changes to minimize energy consumption and ensure smoothness, defined as
\begin{equation*}
L_{\mathrm{R}}=\sum_{k=0}^N{\left( ||\boldsymbol{u}_k||_{\mathrm{R}_{\mathrm{u}}}^{2}+||\dot{\boldsymbol{q}}_{\mathrm{r},k}||_{\mathrm{R}_{\mathrm{q}}}^{2}+||v_{\mathrm{l},k}||_{\mathrm{R}_{\mathrm{v}}}^{2} \right)},
\end{equation*}
where $||\boldsymbol{u}_k||_{\mathrm{R}_{\mathrm{u}}}^{2}=\boldsymbol{u}_{k}^{\top}\mathrm{R}_{\mathrm{u}}\boldsymbol{u}_k$ and similar expressions for other terms. $\mathrm{R}_{\mathrm{u}}$, $\mathrm{R}_{\mathrm{q}}$, and $\mathrm{R}_{\mathrm{v}}$ are positive definite diagonal matrices.

The omnidirectional anisotropy cost $L_{\mathrm{O}}$, inspired by \cite{zhang2024agile}, promotes high-speed and stable movement for the quadruped robot, is defined as
\begin{equation*}
	L_{\mathrm{O}}=\sum_{k=0}^N{\left( \begin{array}{c}
		||\boldsymbol{e}\left( \theta _{\mathrm{r},k} \right) \times \left[ \dot{x}_{\mathrm{r},k},\dot{y}_{\mathrm{r},k} \right] ^{\top}||_{\mathrm{Q}_{\mathrm{\theta}}}^{2}+\\
		||R\left( \theta _{\mathrm{r},k} \right) \left[ \dot{x}_{\mathrm{r},k},\dot{y}_{\mathrm{r},k} \right] ^{\top}||_{\mathrm{Q}_{\mathrm{o}}}^{2} -1
	\end{array} \right)},
\end{equation*}
where $R\left( \cdot \right)$ is rotation matrix and $\mathrm{Q}_{\mathrm{o}}=\mathrm{diag}\left( \left[ v_{\mathrm{xb}}^{-2},v_{\mathrm{yb}}^{-2} \right] \right)$ with the maximum longitudinal velocity $v_{\mathrm{xb}}$ and the maximum lateral velocity $v_{\mathrm{yb}}$. The first term encourages linear velocity in the direction of the robot's head, while the second penalizes states outside the feasible velocity ellipse. 

\subsubsection{Hybrid State Constraint}
We define the hybrid state variable $\boldsymbol{\eta}$ equal to the state transitions of the coarse trajectory $\varGamma$, and formulate hybrid dynamics constraints as \eqref{equ:hybridstate}. The state transition arises from forward dynamics iteration, ensuring natural state switching. This eliminates the need for integer optimization variables, greatly improving solving efficiency. Notably, this approach leverages the mode-switching property of the CT-QR system, allowing us to focus on continuous dynamics while respecting the system's hybrid nature. 
\subsubsection{Input and State Constraint}
The input $\boldsymbol{u}$ and state $\boldsymbol{x}$ are bounded through \eqref{equ:inequalbound}. The input boundaries include the maximum linear acceleration $a_{\mathrm{max}}$ and the maximum angular acceleration $g_{\mathrm{max}}$, defined as
$$
||\left[ \ddot{x}_{\mathrm{r},k},\ddot{y}_{\mathrm{r},k} \right] ||^2\leqslant a_{\max}^{2},\; -g_{\max}\leqslant \ddot{\theta}_{\mathrm{r},k}\leqslant g_{\max}.
$$
State limitations consist of the maximum tractor linear velocity $v_{\mathrm{max}}$ and the maximum tractor linear velocity $\omega _{\mathrm{max}}$, defined as
$$
||\left[ \dot{x}_{\mathrm{r},k},\dot{y}_{\mathrm{r},k} \right] ||^2\leqslant v_{\max}^{2},\; -\omega _{\max}\leqslant \dot{\theta}_{\mathrm{r},k}\leqslant \omega _{\max}.
$$
Trailer is driven by the tractor, thus the trailer velocity and acceleration are within the same range as the tractor, as
$$
0\leqslant v_{\mathrm{l},k}\leqslant v_{\max},\; -\mu G\leqslant {{F_k}/{m_{\mathrm{l}}}}-\mu G\cdot \mathrm{sgn} \left( v_{\mathrm{l},k} \right) \leqslant a_{\max}
$$
where if the trailer's mode switches from slack ($\eta_{k-1} = 0$) to taut ($\eta_{k} = 1$), we set the acceleration constraint as $ 0 \leq {F}_{k}/{m}_{\mathrm{l}} - {\mu G} \cdot  \mathrm{sgn}\left( {v}_{\mathrm{l},k}\right)$ to fit acute change of $v_{\mathrm{l},k}$. Moreover, the steering angle is in the range of $-\Phi \leqslant\delta _k \leqslant \Phi$.
\begin{figure}[htbp]
	\centering
	\includegraphics[width=7.0 cm]{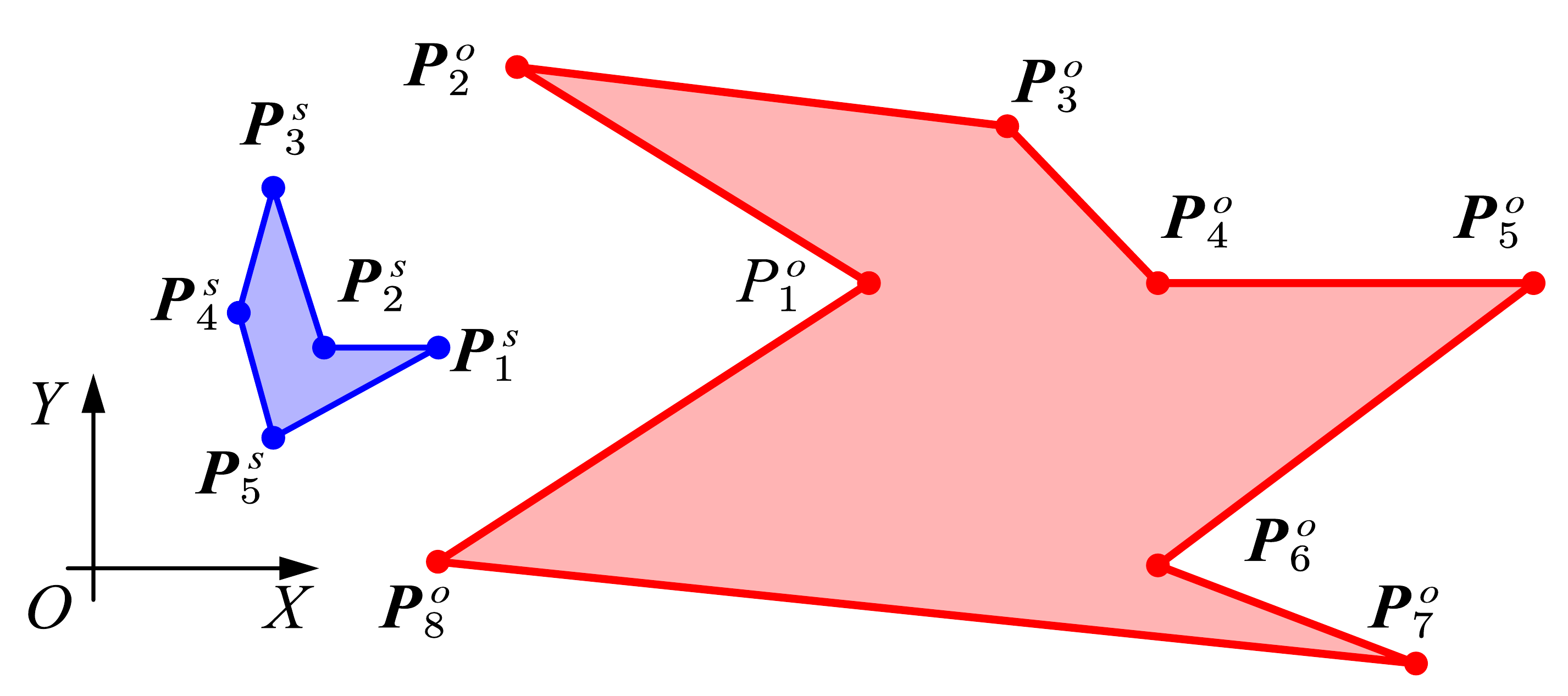}
	\caption{Illustration of the object model of geometry collision avoidance constraints. $\left\{ \boldsymbol{P}_{1}^{\mathrm{s}},\boldsymbol{P}_{2}^{\mathrm{s}}... \right\} \in \mathcal{S}$ represents the system's polygon vertex set and $\left\{ \boldsymbol{P}_{1}^{\mathrm{o}},\boldsymbol{P}_{2}^{\mathrm{o}}... \right\} \in \mathcal{O}$ represents the obstacle's polygon vertex set, where $\boldsymbol{P} = [x,y]$.}
	\label{fig:collisionConfiguration}
\end{figure}
\subsubsection{Collision Avoidance}
The collision avoidance constraints are included in \eqref{equ:obstacle}. For whole body collision avoidance, we model objects as polygons (defined by a set of vertices) as shown in Fig. \ref{fig:collisionConfiguration}. The collision is defined as the edges of two polygons having intersections. To further illustrate, we describe an edge collision avoidance constraint between two edges $l_{1}^{\mathrm{s}}$ and $l_{1}^{\mathrm{o}}$, where each edge can be represented as a convex set formed by two vertices and a parameter as 
\begin{equation*}
	\begin{split}
		l_{1}^{\mathrm{s}}:&=\theta _{1}^{\mathrm{s}}\cdot\, \boldsymbol{P}_{1}^{\mathrm{s}}+\left( 1-\theta _{1}^{\mathrm{s}} \right) \cdot \boldsymbol{P}_{2}^{\mathrm{s}},\;\,\theta _{1}^{\mathrm{s}}\in \left[ 0,1 \right] ,\;\left\{ \boldsymbol{P}_{1}^{\mathrm{s}},\boldsymbol{P}_{2}^{\mathrm{s}}... \right\} \in \mathcal{S} 
		\\
		l_{1}^{\mathrm{o}}:&=\theta _{1}^{\mathrm{o}}\cdot \boldsymbol{P}_{1}^{\mathrm{o}}+\left( 1-\theta _{1}^{\mathrm{o}} \right) \cdot \boldsymbol{P}_{2}^{\mathrm{o}},\;\theta _{1}^{\mathrm{o}}\in \left[ 0,1 \right] ,\;\left\{ \boldsymbol{P}_{1}^{\mathrm{o}},\boldsymbol{P}_{2}^{\mathrm{o}}... \right\} \in \mathcal{O}. 		
	\end{split}
\end{equation*}
\begin{figure}[htbp] 
	\vspace{-0.6cm}
	\centering
	\includegraphics[width=7.0 cm]{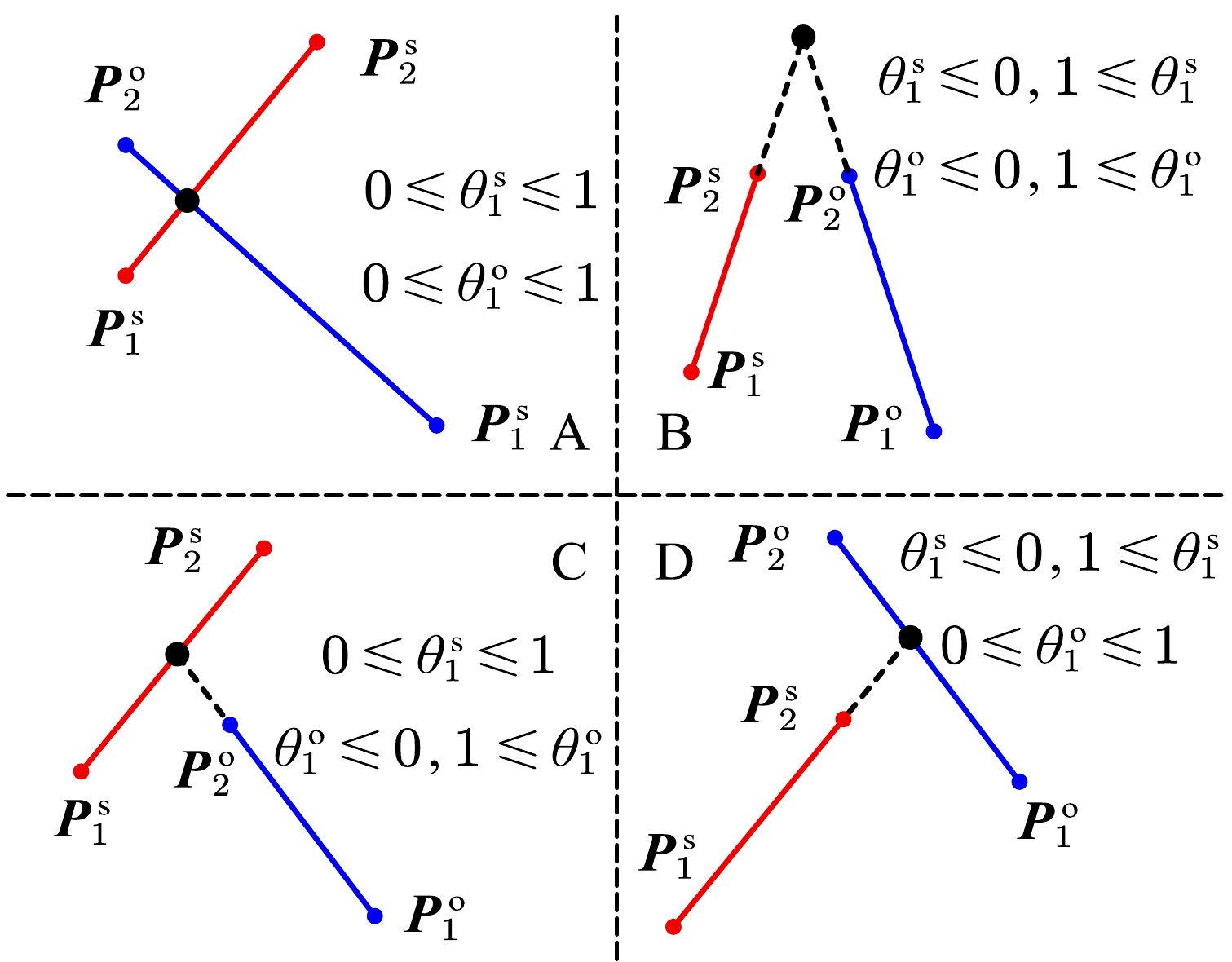}
	\caption{Illustration of four edge conditions: (A) shows collision, while (B), (C), and (D) depict collision avoidance.}
	\label{fig:collisionconditions}
	\vspace{-0.2cm}
\end{figure}

The edge conditions with varying $\theta _{1}^{\mathrm{s}}$ and $\theta _{1}^{\mathrm{o}}$ ranges are shown in Fig. \ref{fig:collisionconditions}.  The virtual intersection point of $l_{1}^{\mathrm{s}}$ and $l_{1}^{\mathrm{o}}$ can be described by vertices as:
\begin{equation*}
	\theta _{1}^{\mathrm{s}}=f_{\uptheta}\left( \boldsymbol{P}_{1}^{\mathrm{s}},\boldsymbol{P}_{2}^{\mathrm{s}},\boldsymbol{P}_{1}^{\mathrm{o}},\boldsymbol{P}_{2}^{\mathrm{o}} \right), \;
	\theta _{1}^{\mathrm{o}}=g_{\uptheta}\left( \boldsymbol{P}_{1}^{\mathrm{s}},\boldsymbol{P}_{2}^{\mathrm{s}},\boldsymbol{P}_{1}^{\mathrm{o}},\boldsymbol{P}_{2}^{\mathrm{o}} \right). 
\end{equation*}
Additionally, parallel edges belong to case Fig. \ref{fig:collisionconditions}(B), where $\theta _{1}^{\mathrm{s}}=\infty$ and $\theta _{1}^{\mathrm{o}}=\infty$. 

The collision-free conditions for $l_{1}^{\mathrm{s}}$ and $l_{1}^{\mathrm{o}}$ can be formulated as $\theta _{1}^{\mathrm{s}}\notin \left[ 0,1 \right] \cup \theta _{1}^{\mathrm{o}}\notin \left[ 0,1 \right]$. Furthermore, it can be expressed as an inequality as 
\begin{equation}
	\left\| \left[ \theta _{1}^{\mathrm{s}}-0.5,\theta _{1}^{\mathrm{o}}-0.5 \right] \right\| _{\infty}\geqslant 0.5. \label{equ:infinitenorm}
\end{equation}

However, using the infinite norm increases the complexity of the optimization. To improve numerical stability, we further scale the bounds, resulting in 
\begin{equation}
	\left( \theta _{1}^{\mathrm{s}}-0.5 \right) ^2+\left( \theta _{1}^{\mathrm{o}}-0.5 \right) ^2\geqslant 0.5. \label{equ:scalarinequal}
\end{equation}

\begin{figure}[htbp] 
	\vspace{-0.4cm}
	\centering
	\includegraphics[width=8.4 cm]{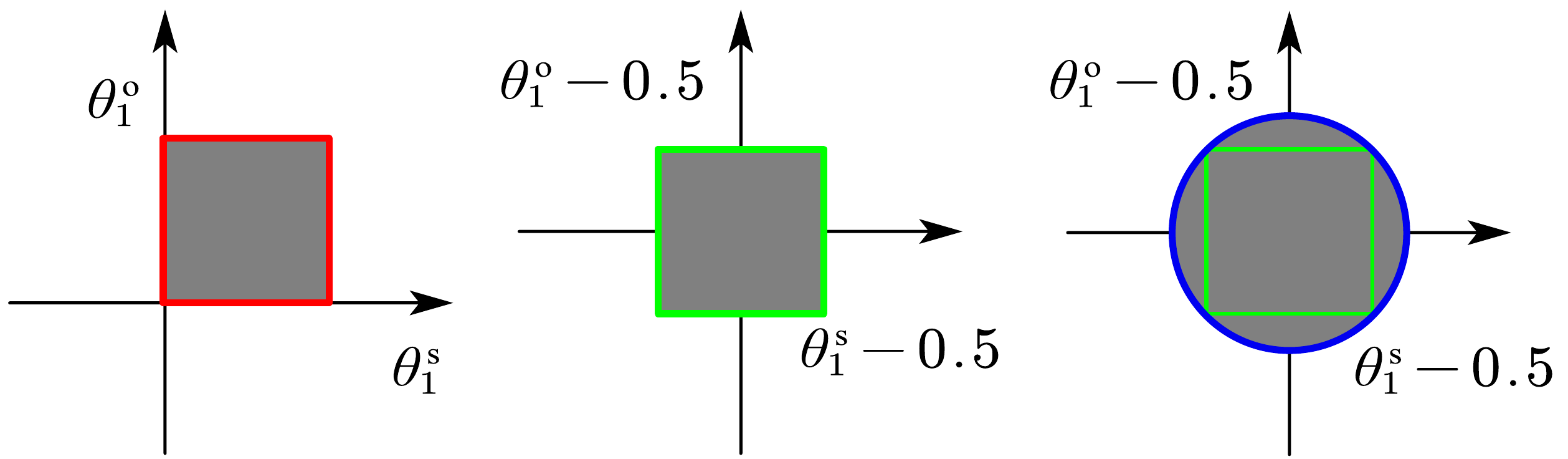}
	\caption{Red square illustrates the original collision avoidance region. Green square illustrates infinite norm region \eqref{equ:infinitenorm}. Blue circle denotes scaled inequality constraint \eqref{equ:scalarinequal}.}
	\label{fig:thetarange}
\end{figure}

Aforementioned collision avoidance regions are shown in Fig. \ref{fig:thetarange} for clarity. It is important to highlight that our collision avoidance constraint is applicable to all variable polygons and can be easily extended to higher-dimensional spaces.

\begin{figure*}[bp] 
	\vspace{-0.2cm}
	\centering
	\subfigure[Proposed-TO]
	{
		\begin{minipage}[b]{.17\linewidth}
			\centering
			\includegraphics[height=2.4 cm]{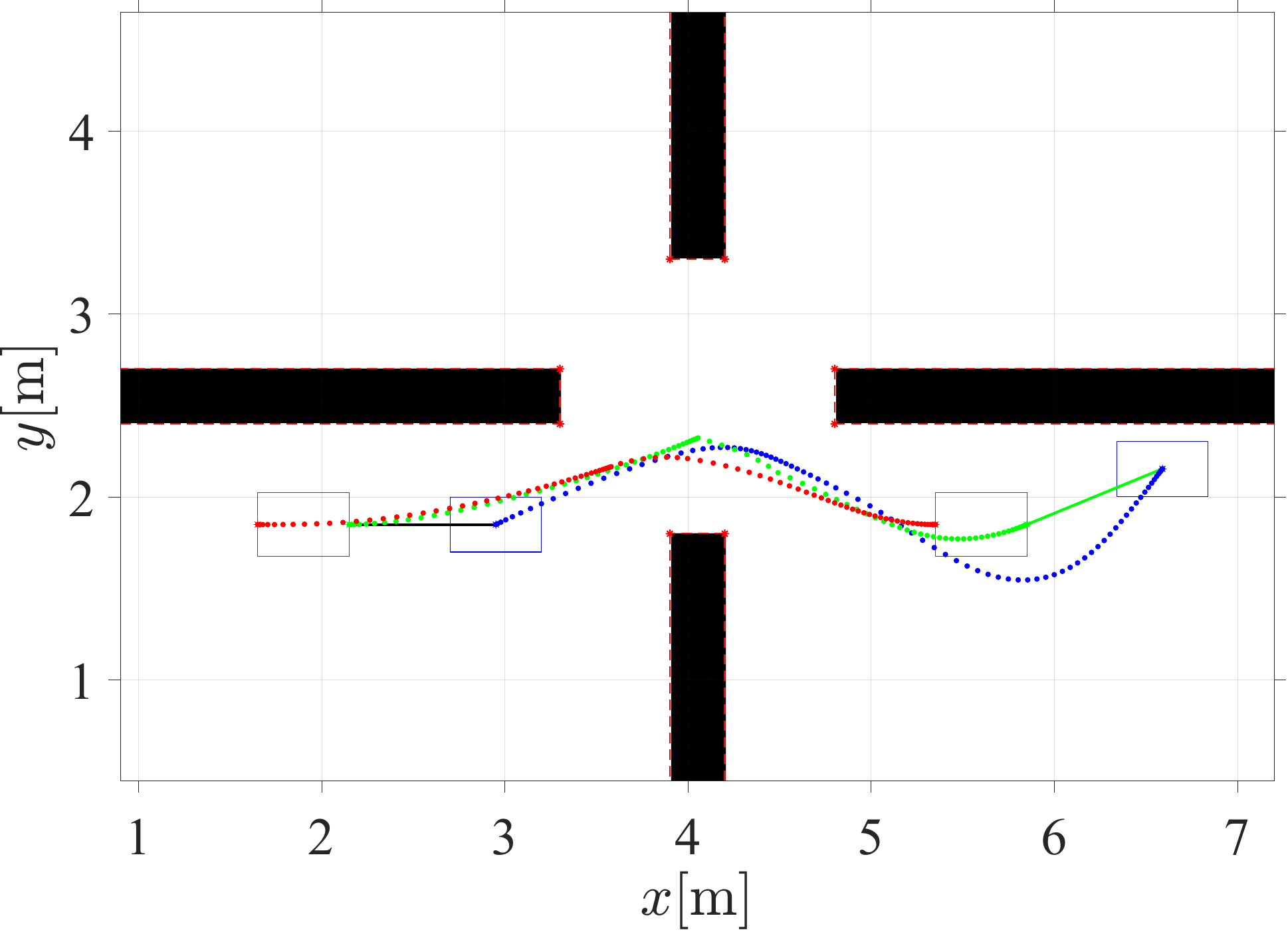}
			\\
			\includegraphics[height=2.4 cm]{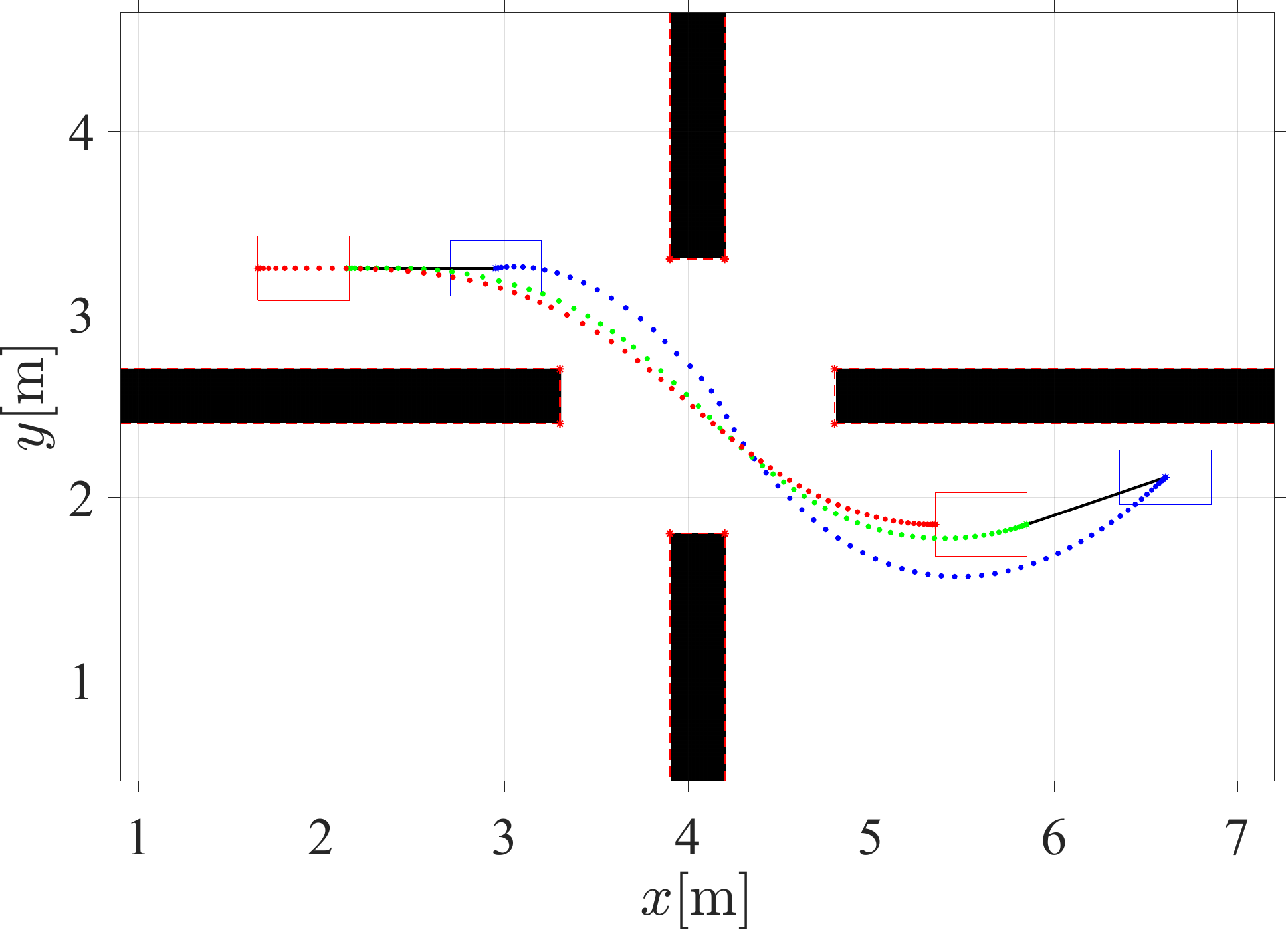}
			\\
			\includegraphics[height=2.4 cm]{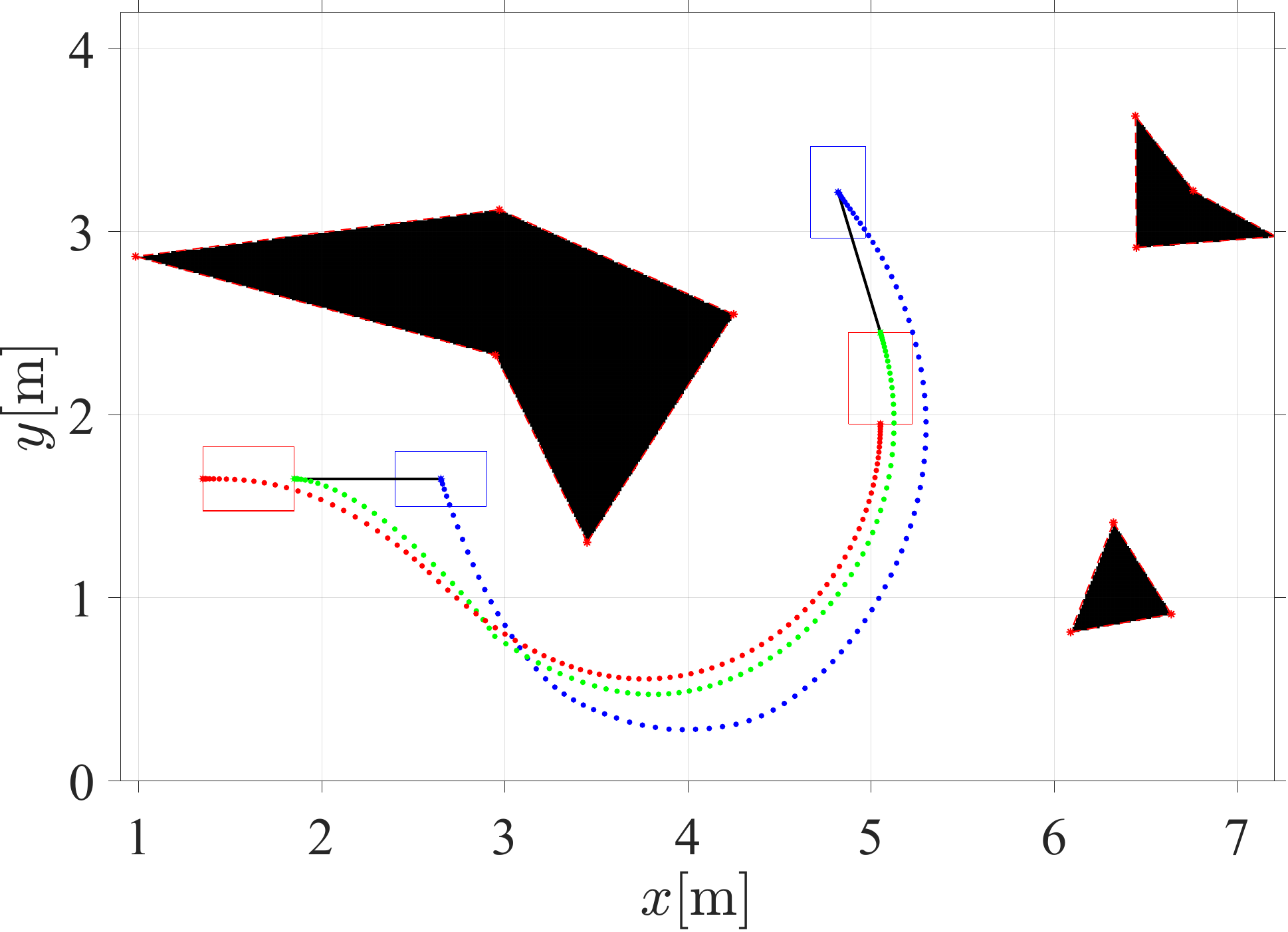}
		\end{minipage}
		\label{fig:optimalgeneral}
	}
	\subfigure[MINLP-TO*]
	{
		\begin{minipage}[b]{.17\linewidth}
			\centering
			\includegraphics[height=2.4 cm]{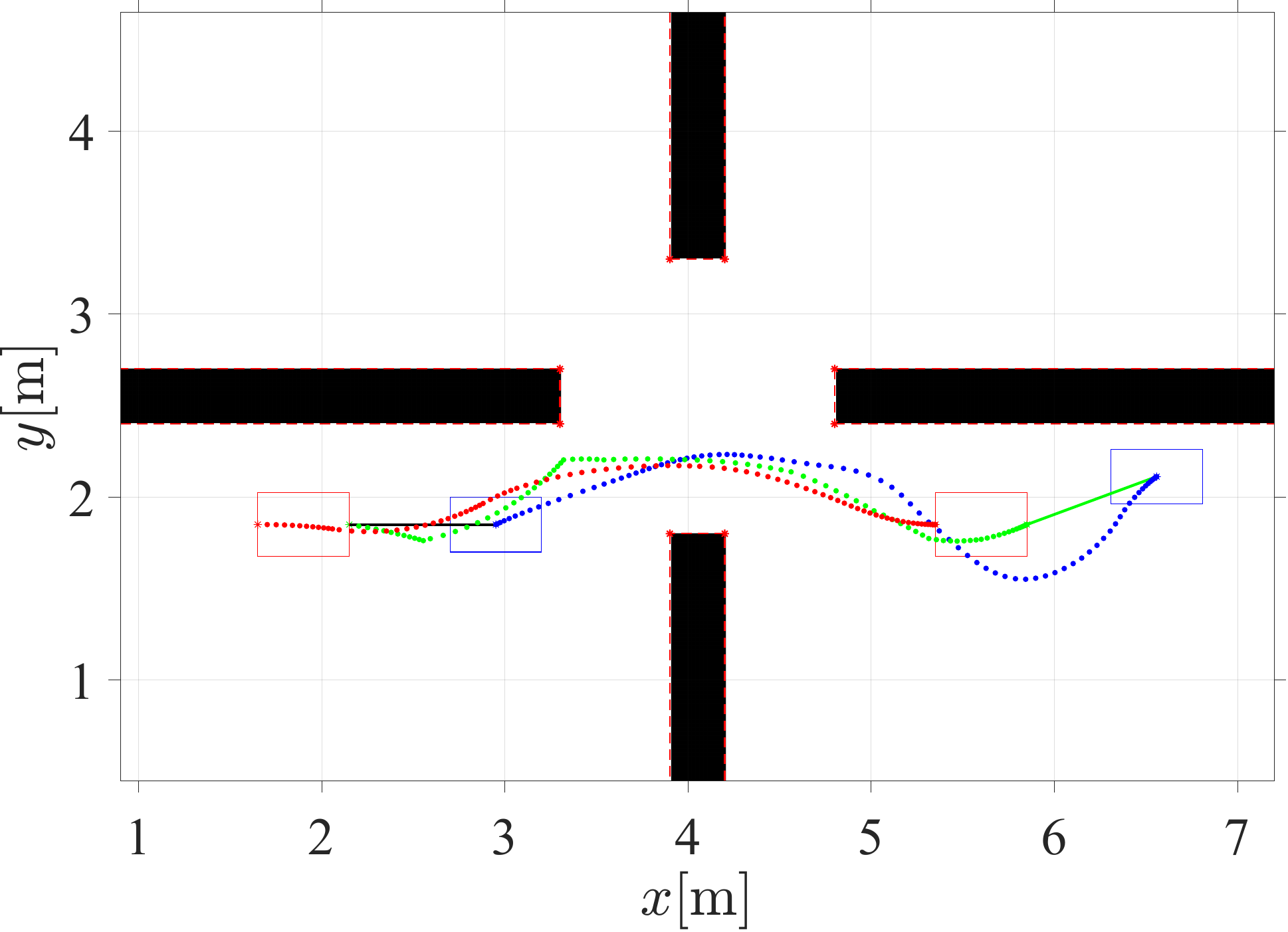}
			\\
			\includegraphics[height=2.4 cm]{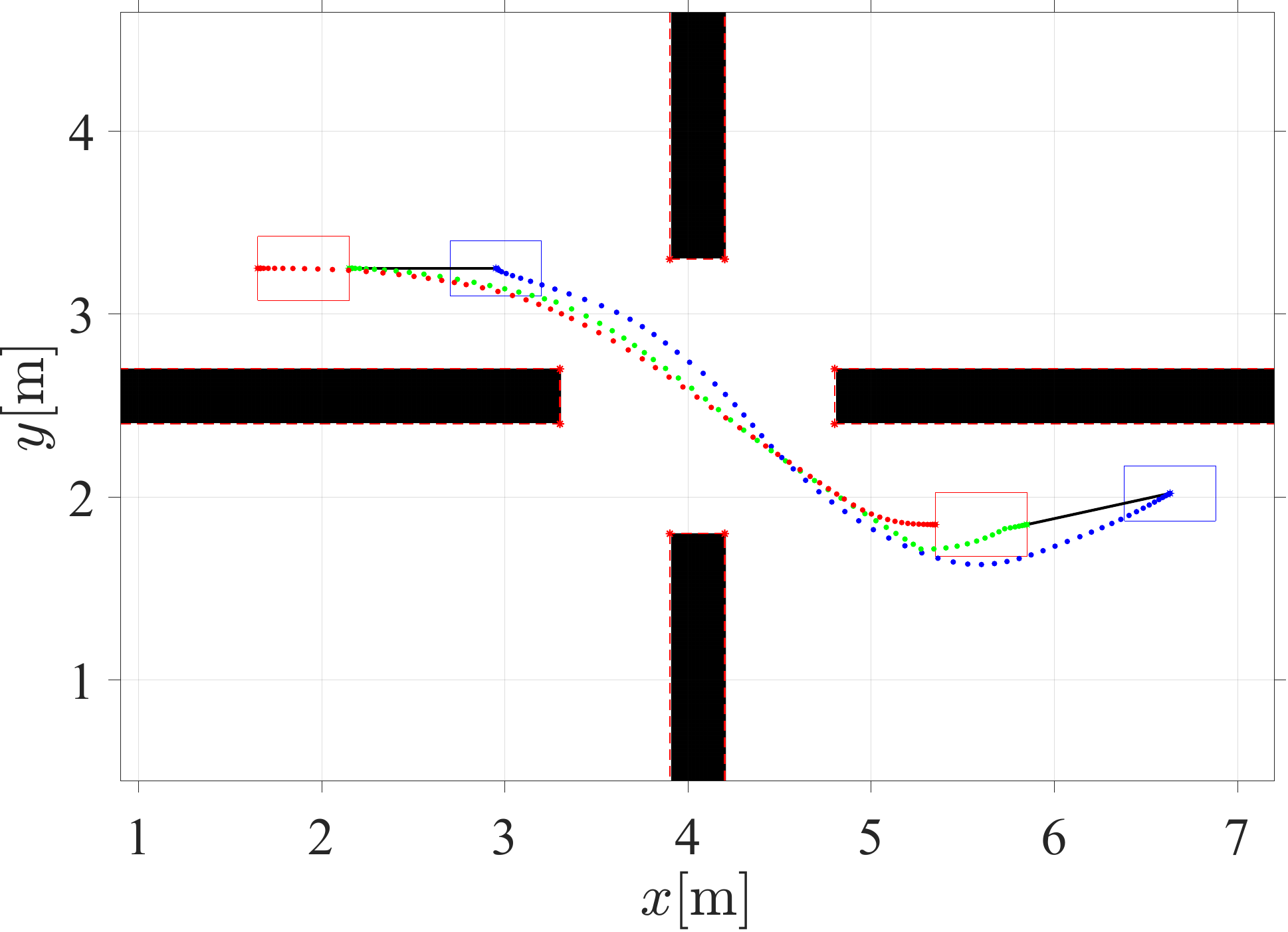}
			\\
			\includegraphics[height=2.4 cm]{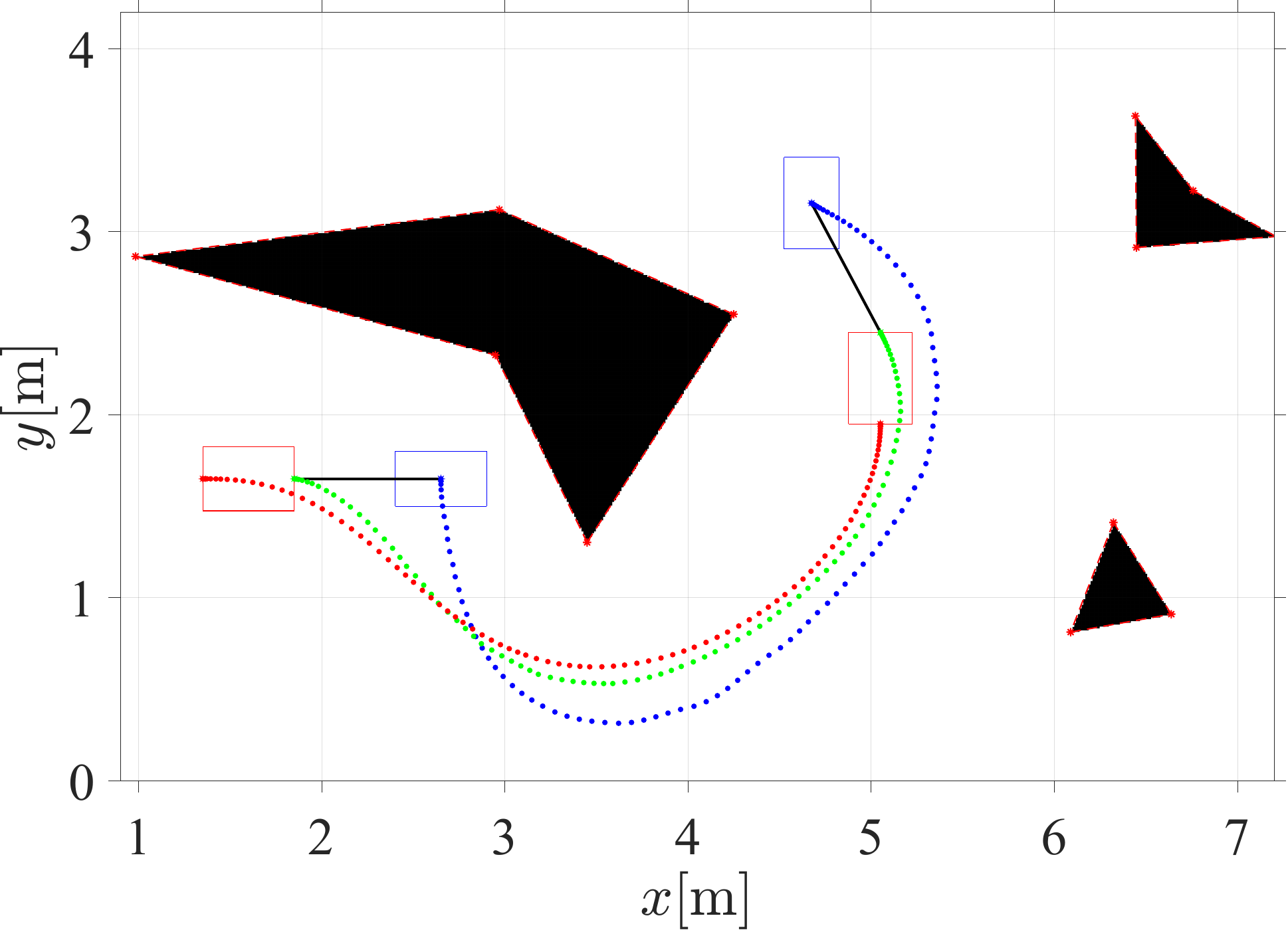}
		\end{minipage}
		\label{fig:minlpgeneral}
	}
	\subfigure[Tension-TO]
	{
		\begin{minipage}[b]{.17\linewidth}
			\centering
			\includegraphics[height=2.4 cm]{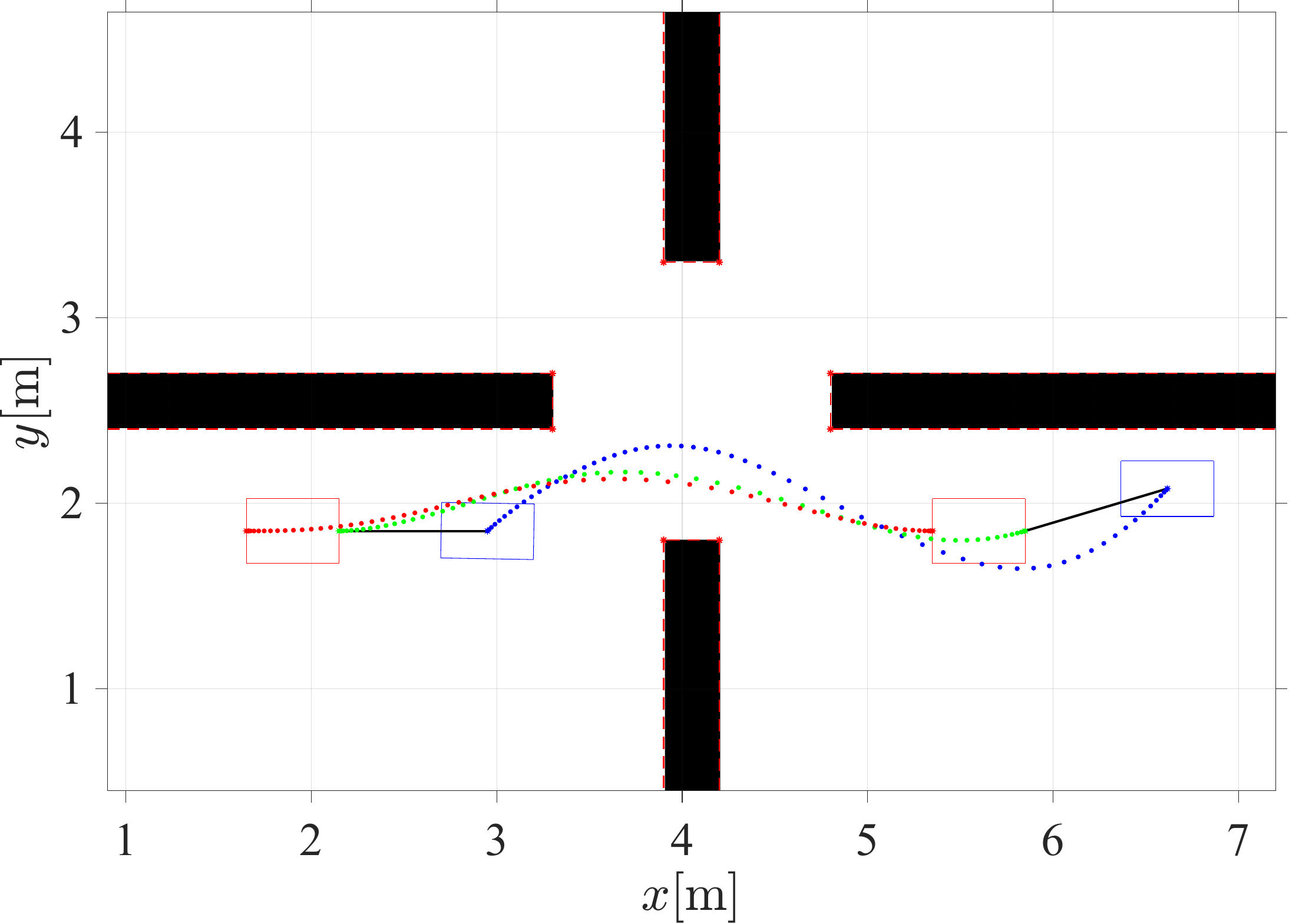}\\
			\includegraphics[height=2.4 cm]{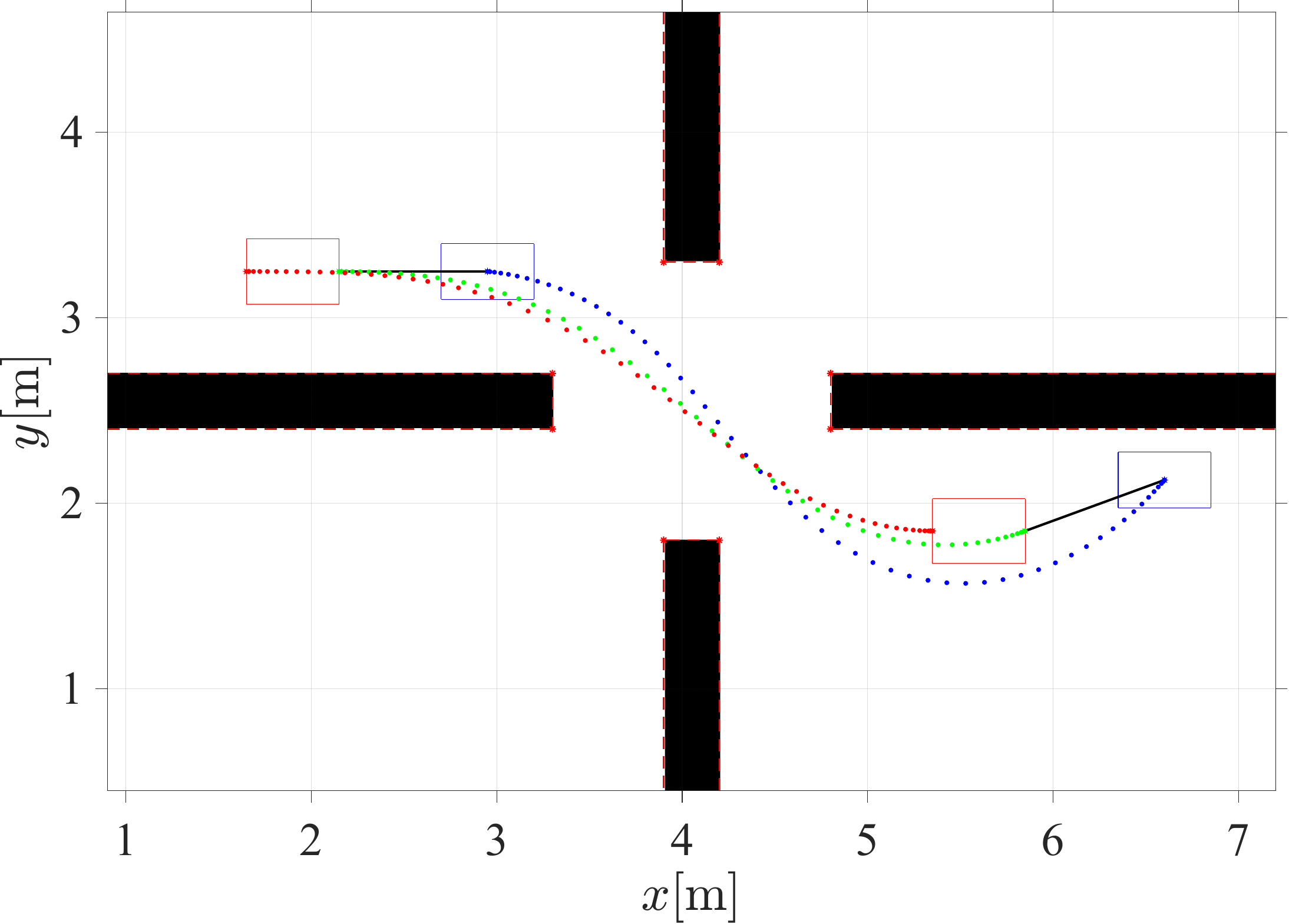}
			\\
			\includegraphics[height=2.4 cm]{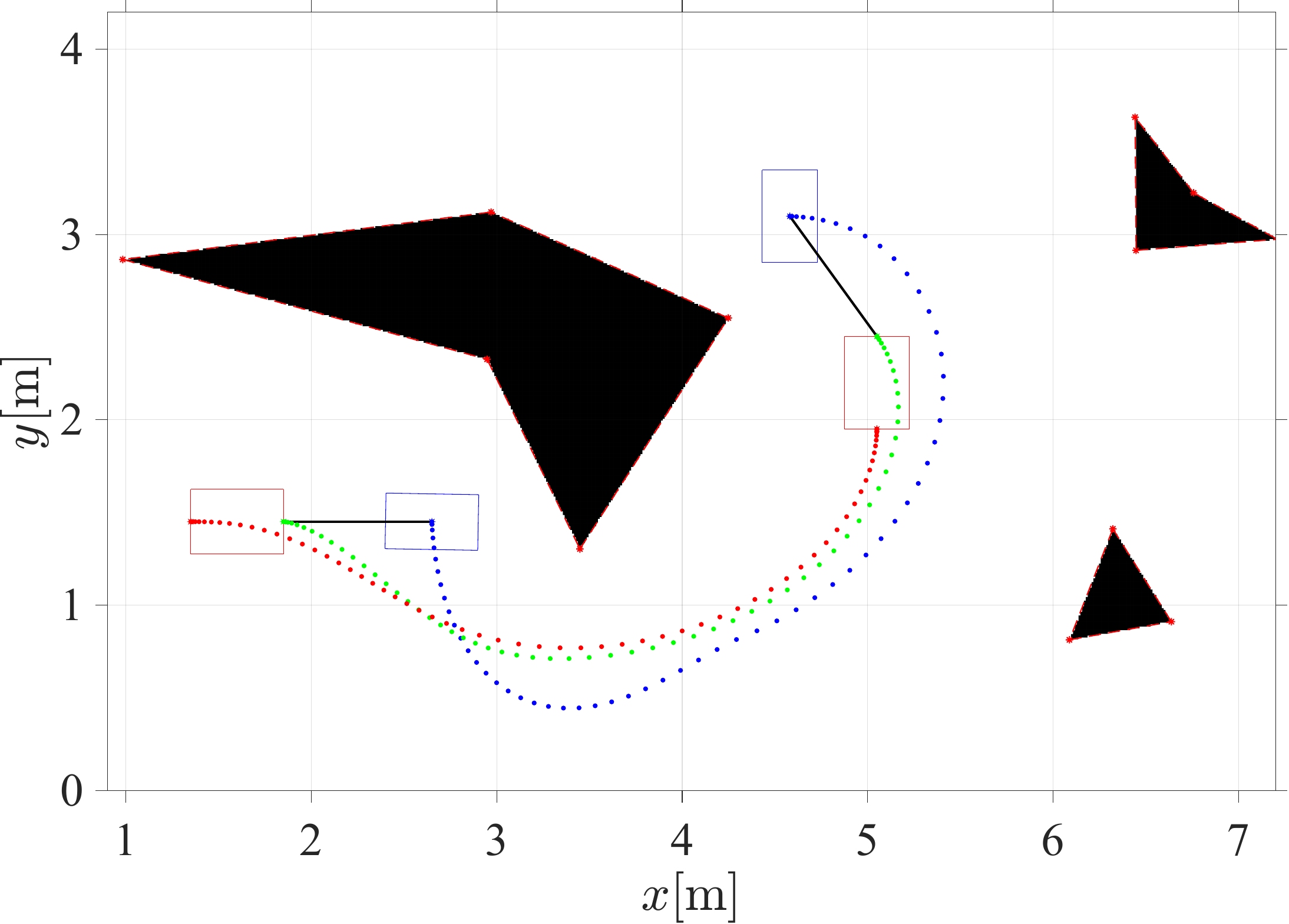}
		\end{minipage}
		\label{fig:searchgeneral}
	}
	\subfigure[MoCap]
	{
		\begin{minipage}[b]{.17\linewidth}
			\centering
			\includegraphics[height=2.4 cm]{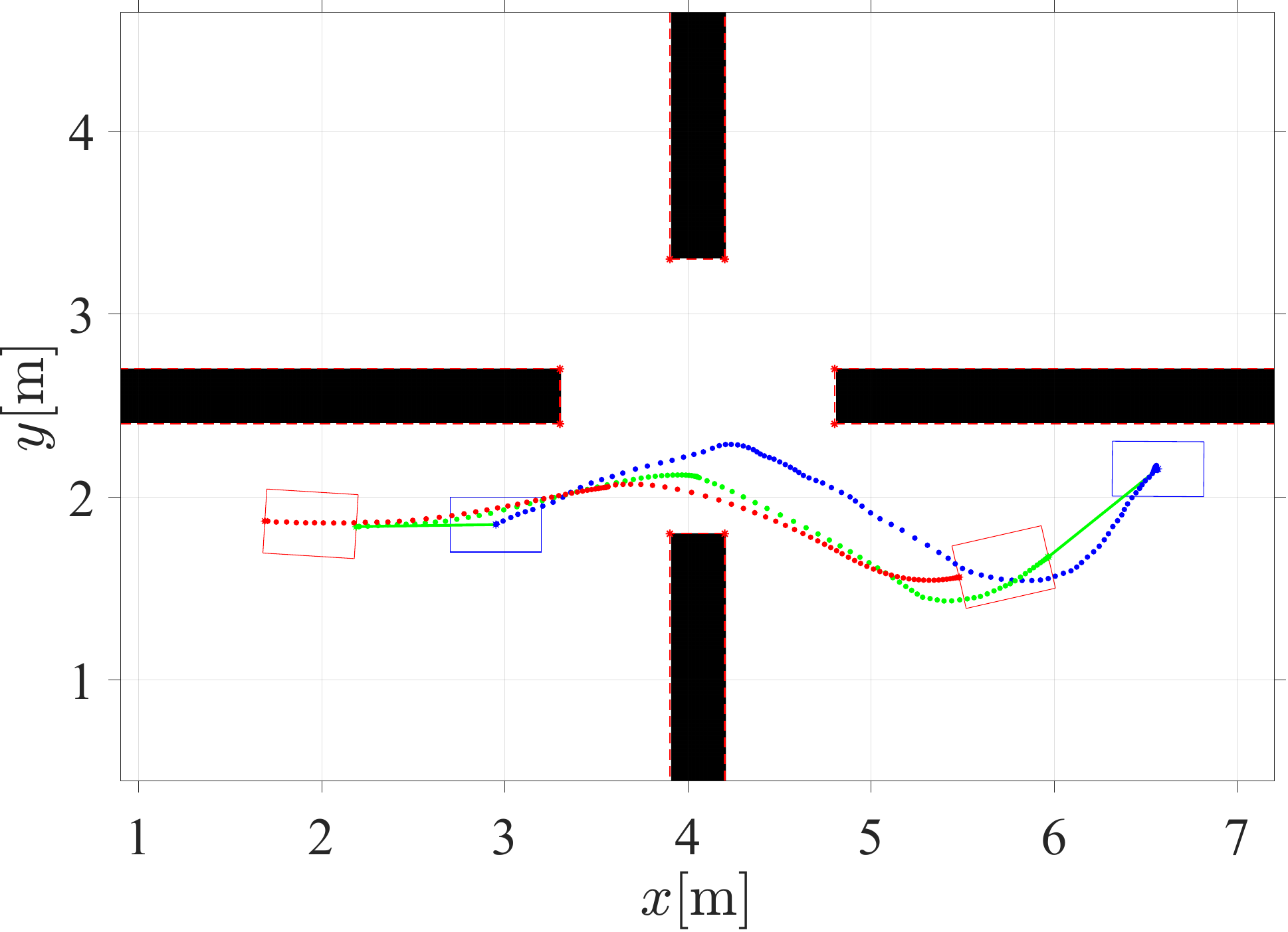}
			\\
			\includegraphics[height=2.4 cm]{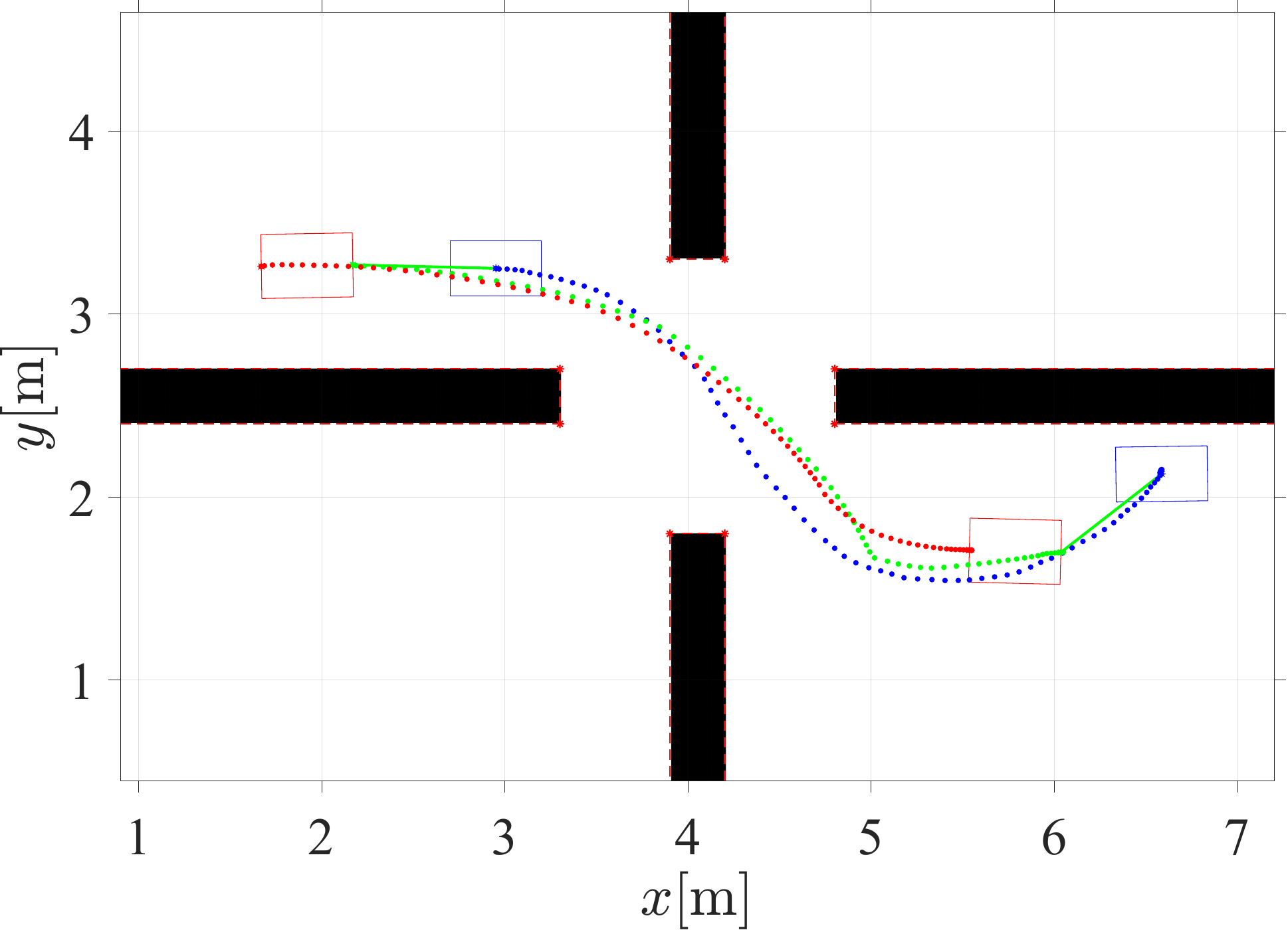}
			\\
			\includegraphics[height=2.4 cm]{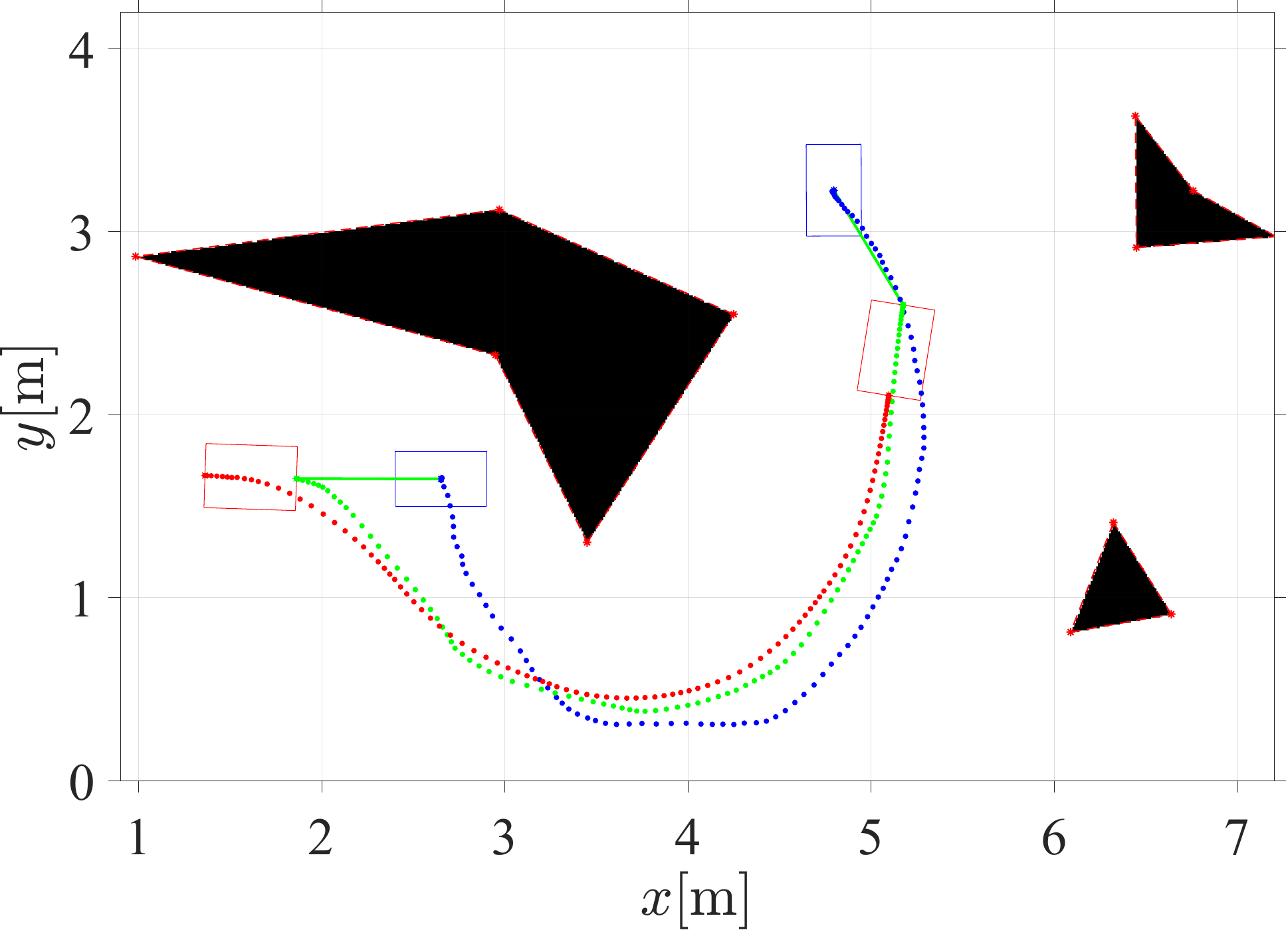}
		\end{minipage}
		\label{fig:nokovgeneral}
	}
	\subfigure[Real-World]
	{
		\begin{minipage}[b]{.22\linewidth}
			\centering
			\raisebox{0.12cm}{\includegraphics[height=2.3 cm]{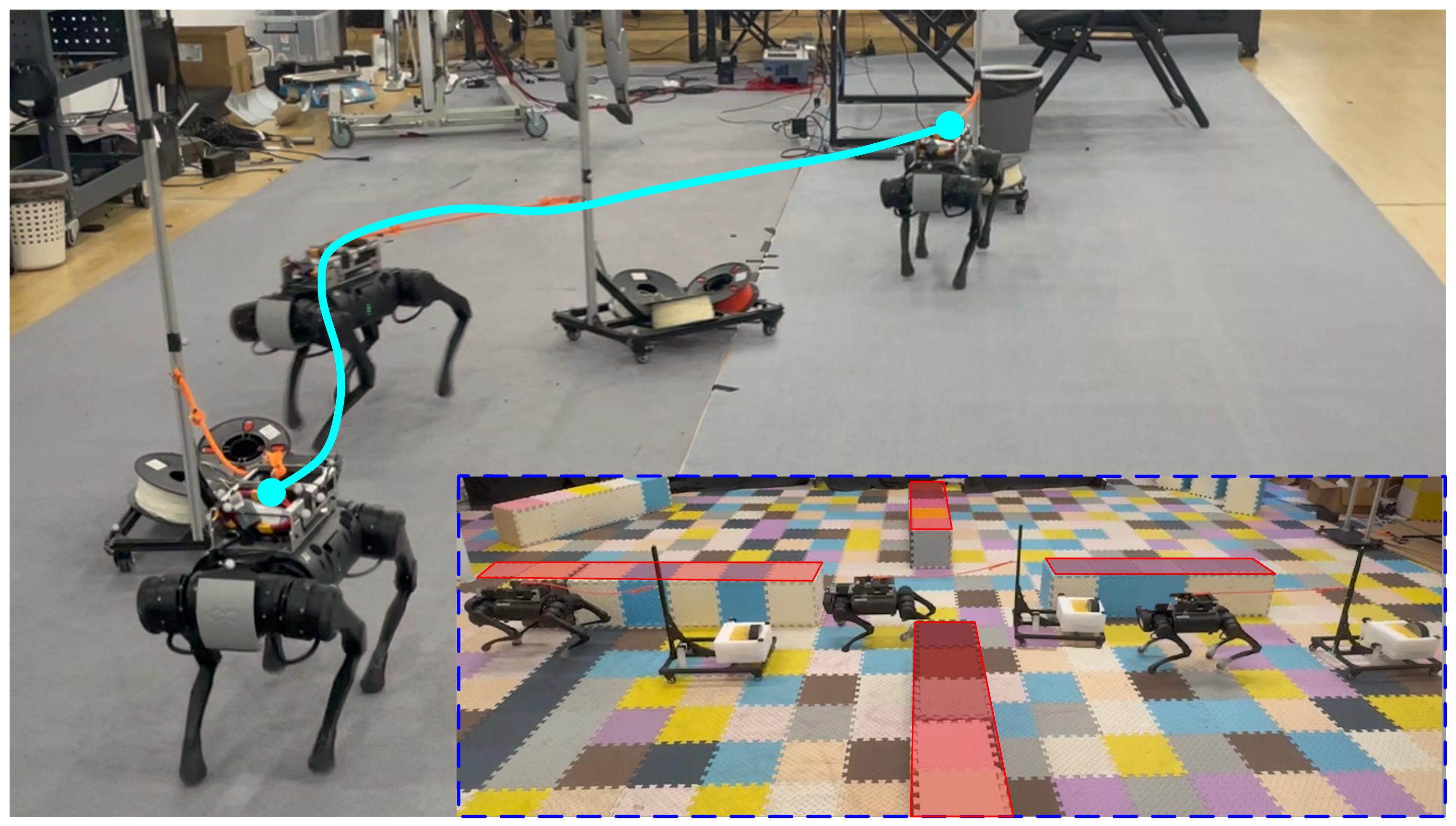}}
			\\
			\raisebox{0.12cm}{\includegraphics[height=2.3 cm]{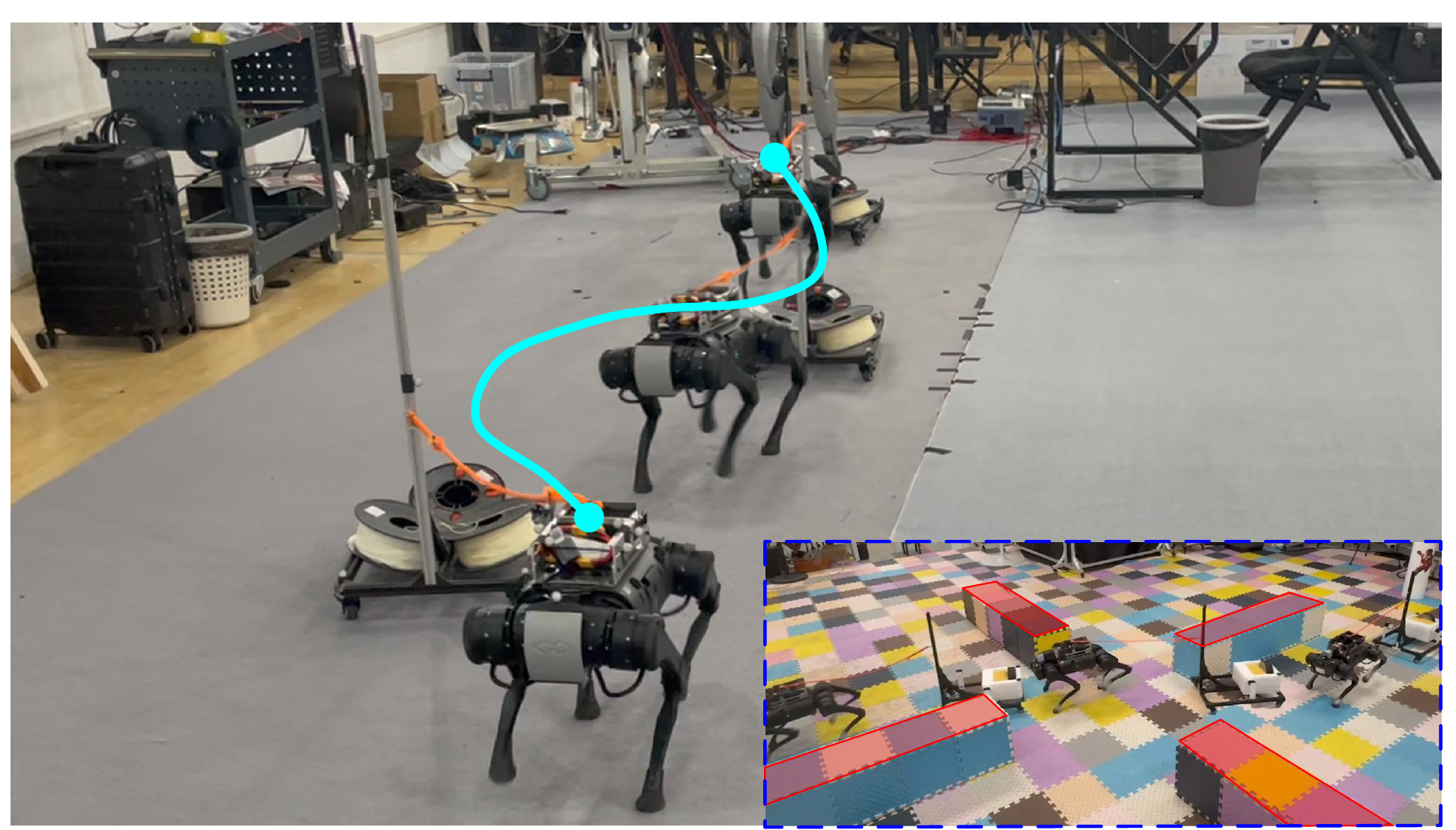}}
			\\
			\raisebox{0.12cm}{\includegraphics[height=2.3 cm]{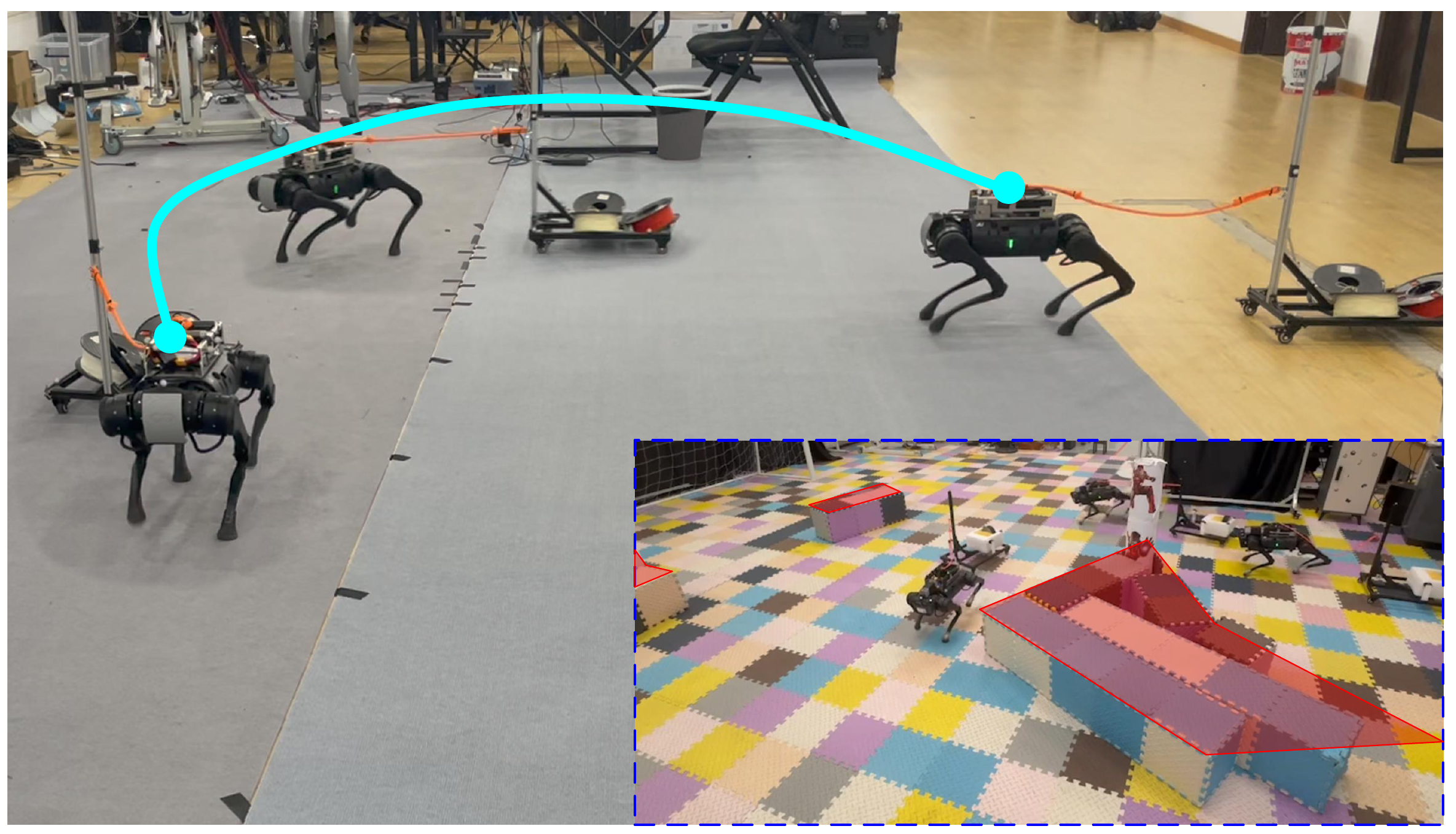}}
		\end{minipage}
		\label{fig:realrobot}
	}
	\caption{Trajectory planning results of the CT-QR system. Each row of figures represents the results of a specific experiment. Blue point presents quadruped robot, green point presents cable, and red point presents trailer. The obstacles are presented by black and the geometric collision boundaries used for collision avoidance constraint are presented by red dashed lines. Cyan lines represent tractor motion trace in the real-world. } 
	\label{fig:indoor}
\end{figure*}

\section{Experiments and Discussion}\label{sec:experiments}
We conduct a series of experiments, both in simulation and in real-world, to validate the proposed hybrid model and planning methods.

\subsection{Experiments Setup}
\subsubsection{Algorithm Implementation} 
The hybrid search and trajectory optimization modules are implemented in MATLAB and run on a desktop computer equipped with an Intel Core i7-12700 CPU and 32 GB of RAM. The trajectory optimization problem is formulated using CasADi \cite{andersson2019casadi}, with IPOPT \cite{biegler2009ipopt} and Bonmin \cite{pierre2023bonmin} acting as solvers for NLP (Nonlinear Programming) and MINLP (Mixed Integer Nonlinear Programming), respectively. The tractor's objective is to drive the trailer to a specified pose while avoiding obstacles. The trajectory is planned offline and then tracked in real-time under 10 Hz on the real CT-QR system.
\subsubsection{System Implementation}
The real CT-QR system is an Unitree A1 quadruped robot equipped with a customized cable-trailer, as shown in Fig. \ref{fig:robotsled}. The NMPC-WBC controller \cite{liao2023walking} handles robot locomotion control and trajectory tracking by feeding a time-sequence-based optimal state trajectory of tractor's body as the reference. The general formulation of the NMPC problem is based on \cite{grandia2023perceptive}. Indoor localization is provided by the NOKOV motion capture system, while outdoor localization relies on LiDAR-Inertial Odometry \cite{xu2022fast}. 

\subsection{General Experiments}
We perform three general experiments, comparing different scenarios to validate the effectiveness and efficiency of the proposed methods, as well as to analyze the quality of the planned trajectories. The details of the algorithm parameters are provided in Table \ref{tab:parameters}.

\begin{table}[htbp]
	\renewcommand{\arraystretch}{1.6}
	\centering
	\caption{ALGORITHM PARAMETERS}\label{tab:parameters}
	\setlength{\tabcolsep}{1pt} 
	\resizebox{8.6cm}{!}{
	\begin{tabular}{ccccc} 
	\hline
	$a_{\mathrm{max}}$=\SI{1.0}{m/s^2}   &$g_{\mathrm{max}}$=\SI{1.5}{rad/s^2}	&$v_{\mathrm{max}}$=\SI{1.0}{m/s}	&$\omega_{\mathrm{max}}$=\SI{1.5}{rad/s}   &$T_{\mathrm{s}}$=\SI{0.5}{s}  \\ 
	\hline
	$D_{\mathrm{a}}$=\SI{0.25}{m/s^2}  & $D_{\uptheta}$ =$\pi/12$\,\SI{}{rad}	& $D_{\mathrm{d}}$ =\SI{0.20}{m}	& $D_{\mathrm{r}}$ =$\pi/12$\,\SI{}{rad}   & $\mathrm{d}t$ =\SI{0.1}{s}  \\ 
	\hline
	\end{tabular}
	}
\end{table}

\subsubsection{Comparative Study}
The trajectory indices for different scenarios are presented in Fig.~\ref{fig:indoor} and Table~\ref{tab:comparison}. Here, HS refers to searching trajectories, while TO indicates trajectory optimization with HS as the initial guess. To further evaluate the efficiency of our planning method, we model the state transition as integer variables and apply MINLP for trajectory optimization, TO* represents the optimization process halted once the first feasible solution of MINLP is found. While, we also compare with tension-only planning, which is widely used in quadrotor suspended-payload problems. ``Out'' signifies that the solving process was terminated after exceeding 24 hours.

\begin{table}[htbp]
	\renewcommand{\arraystretch}{1.4}
	\setlength\tabcolsep{3pt}
	\centering
	\caption{OPTIMIZATION INDEX}\label{tab:comparison}
	\begin{tabular}{c cc cc cc} 
	\hline
	 & \multicolumn{2}{c}{Proposed} 
	 & \multicolumn{2}{c}{MINLP}
	 & \multicolumn{2}{c}{Tension} \\ 
			\cmidrule(lr){2-3} \cmidrule(lr){4-5} \cmidrule(lr){6-7}
	 & HS & TO & TO* & TO & HS & TO\\
	\hline
	1 &  5.2s/32.1     & 62.7s/24.9	& 4.3ks/27.6	& Out & 24.2s/40.3     & 53.1s/52.3  \\ 
	2 &  9.5s/71.8      & 45.5s/71.8	& 10.1ks/66.4	& Out & 15.6s/59.8     & 62.7s/91.1 \\
	3 &  15.5s/67.9       & 59.2s/93.2 & 1.02ks/112.1	& Out & 18.3s/72.3     & 79.9s/173.9 \\
	\hline
	\end{tabular}
\end{table}

In general, for \textbf{Non-convex} and \textbf{Nonlinear} optimization problems, the initial guess  by HS plays a crucial role in both the convergence and the quality of the solution. TO fails to converge without a suitable initial guess. The TO trajectories are smoother and more energy-efficient than the HS trajectories. Moreover, the MINLP-TO* trajectories closely resemble the Proposed-TO trajectories, but their solving process is significantly more time-consuming. Tension-TO may generate longer trajectories due to the absence of slack modes, which can lead to failure in complex environments as discussed in Section~\ref{sec:LshapeExp}.

\begin{figure*}[bp] 
	\centering
	\subfigure[Target]
	{
		\begin{overpic}[width=5.4 cm]{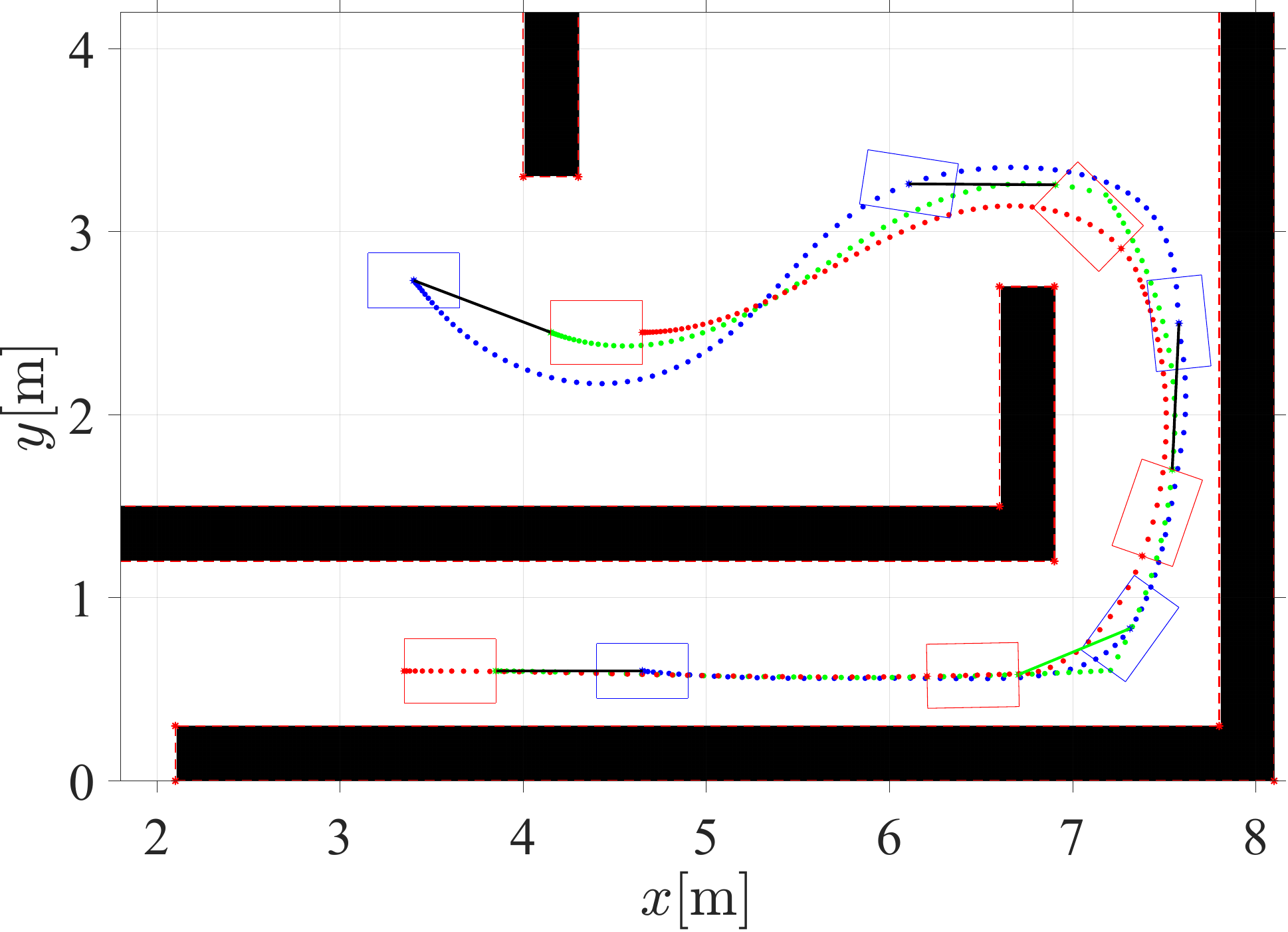}
            \put(40,23){\scriptsize (A)}
			\put(73,24){\scriptsize (B)}
			\put(82,40){\scriptsize (C)}
			\put(72,62){\scriptsize (D)}
			\put(40,50){\scriptsize (E)}
        \end{overpic}
		\label{fig:optimalnarrow}
	}
	\subfigure[Real]
	{
		\begin{overpic}[width=5.4 cm]{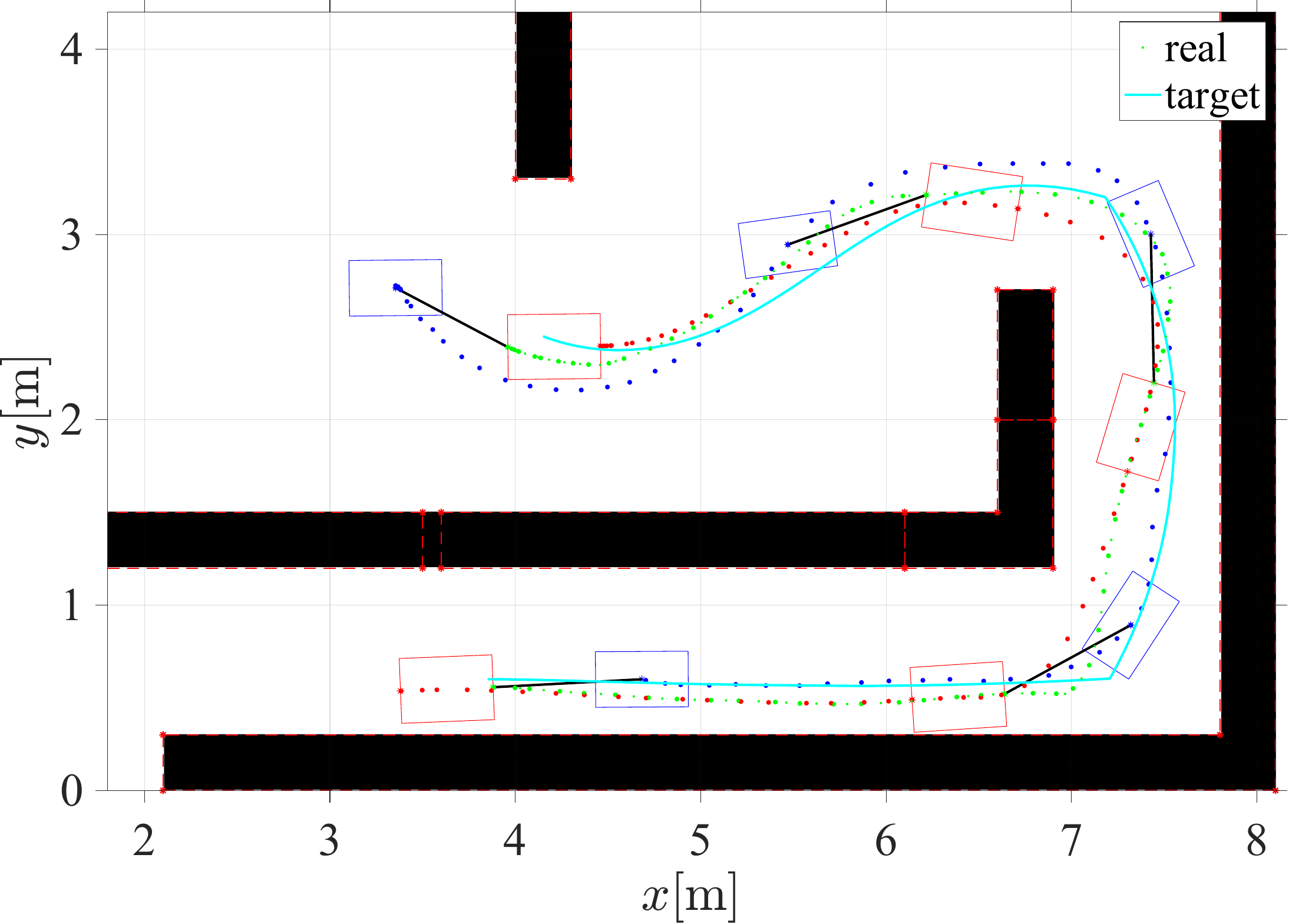}
			\put(40,23){\scriptsize (A)}
			\put(73,24){\scriptsize (B)}
			\put(82,40){\scriptsize (C)}
			\put(72,62){\scriptsize (D)}
			\put(40,50){\scriptsize (E)}
		\end{overpic}
		\label{fig:nokovnarrow}
	}
	\subfigure[Tracking States]
	{
		\includegraphics[width=5.8 cm]{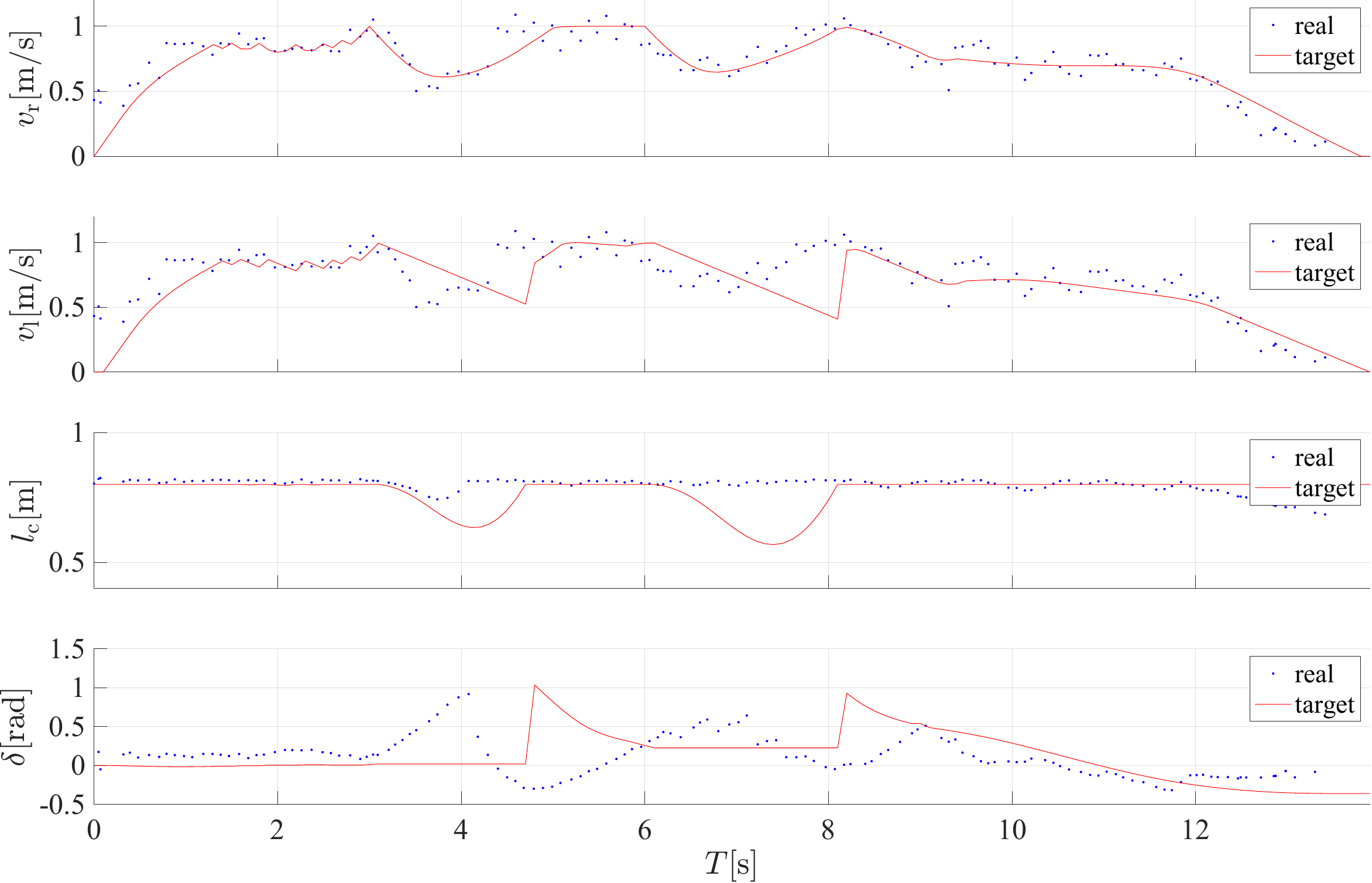}
		\label{fig:statesnarrow}
	}
	\caption{Illustration of challenging trajectory planning in an L-shaped narrow corridor. Cyan line in (b) shows the target trailer trajectory. Initially, the tractor accelerates to \SI{0.8}{m/s} and pulls the trailer to the corner. Then, the tractor decelerates and turns left to avoid colliding with the wall, while the trailer decelerates freely and reaches the corner (A-B). After that, the tractor accelerates again and pulls the trailer to turn left to pass the corner (C-E). Finally, the tractor decelerates along with the trailer to stop at the desired pose (E)} 
	\label{fig:LshapeHSTO}
\end{figure*}

\subsubsection{Discussion}
The motion capture system (MoCap) results and real-world results are shown in Fig. \ref{fig:nokovgeneral} and Fig. \ref{fig:realrobot}, while planning indices are detailed in Tab.~\ref{tab:experiment}. Within the \SI{5.1}{m} $\times$ \SI{8.1}{m} area, the robot successfully maneuvers the trailer to the desired pose, achieving distance errors within \SI{0.23}{m} and angle errors within \SI{0.22}{rad} at a speed close to \SI{1}{m/s}. Exp 3* refers to trajectory tracking at half velocity, which results in significantly smaller trailer errors. This is likely because, at lower speeds, the trailer's deceleration upon reaching the goal pose reduces movement distance errors, minimizing the impact of uneven ground friction coefficients.

\begin{table}[htbp]
	\vspace{0.2cm}
	\renewcommand{\arraystretch}{1.4}
	\setlength\tabcolsep{3pt}
	\centering
	\caption{EXPERIMENT RESULT}\label{tab:experiment}
	\begin{tabular}{lccccc} 
	\hline
	& $ q_{\mathrm{l},\mathrm{goal}}$ error  & $q_{\mathrm{l},k}$ error& $q_{\mathrm{r},k}$ error& $v_{\mathrm{r},k}$ & $v_{\mathrm{r},k}$ \\ 
	 & [m]/[rad] & [m]/[rad](ave) & [m]/[rad](ave) & [m/s](ave) & [m/s](max) \\
	\hline
	1   	&0.22/0.22   &0.18/0.10	&\textbf{0.02/0.01}	& 0.52  & 0.93 \\ 
	2   	&0.23/0.02   &0.13/0.11	&0.02/0.02	& \textbf{0.75}  & \textbf{1.11}  \\ 
	3   	&0.18/0.16   &\textbf{0.09/0.08}	&0.02/0.02	& 0.67  & 0.99  \\ 
	3*		& \textbf{0.02/0.02}   &0.17/0.10	&0.03/0.01	& 0.39  & 0.65  \\ 
	\hline
	\end{tabular}
\end{table}
In various scenarios, the tractor's motion remains stable, and it successfully navigates through obstacles, showcasing the practical application of our obstacle avoidance constraints. The actual motion trace of the trailer closely follows the desired trajectory generated by our planning method, confirming a strong alignment between our proposed hybrid dynamics model and the real system, resulting in small tracking errors, such as 0.09m and 0.08rad. 
The outdoor experiments shown in Fig. \ref{fig:outdoortracking}, further demonstrating the robustness of our planning approach in real-world. Additionally, more details are available in the supplementary video.


\subsection{Narrow Corridor Experiment}\label{sec:LshapeExp}
The CT-QR system is capable of navigating narrow environments, with the trailer being maneuvered by hybrid cable states. To validate this capability, we conduct an experiment in an L-shaped corridor with a width of \SI{0.9}{m}.

\subsubsection{Trajectory Planning}
The planned trajectory is shown in Fig. \ref{fig:optimalnarrow}, the tractor maneuvers the trailer through the narrow corridor, leveraging the cable's taut/slack mode transition. Due to the complex shape of the corridor, the trajectory planning process is time-consuming, taking about \SI{1.46}{ks} to find a feasible trajectory by search module and \SI{121.45}{s} to refine trajectory with optimization module. Notably, under tension-only planning, the search process fails to find a feasible trajectory even after all nodes have been expanded.

\subsubsection{Discussion}
The experimental results are shown in Fig. \ref{fig:nokovnarrow}, Fig. \ref{fig:statesnarrow} and Fig. \ref{fig:narrowexperiment}. The tractor maneuvers the trailer through the narrow corridor with the average speed of \SI{0.76}{m/s}, maximum linear speed of \SI{1.11}{m/s}, and maximum angular speed of \SI{0.26}{rad/s}, demonstrating the effectiveness of our method in leveraging hybrid state transition. The discontinuous motion of the quadruped robot impacts the smoothness of the trailer's movement, and the trailer's deceleration results in varying accelerations (uneven ground friction). Due to these factors, the trailer pose error is \SI{0.18}{m} and \SI{0.0024}{rad}, and the real-world cable length does not match the target trajectory perfectly. Nevertheless, our method successfully guides the CT-QR system to the target pose, proving its effectiveness in planning trajectories for the CT-QR system in challenging environments.

\section{Conclusion}\label{sec:conclusion} %
In this paper, we propose the cable-trailer with quadruped robot system and introduce a hybrid state dynamics model for motion planning. We then present a hierarchical motion planning method: the front-end trajectory generation, inspired by hybrid A* search, incorporates state transitions in forward dynamics iteration, while the back-end trajectory optimization ensures smooth, energy-efficient, safe, and dynamic feasible trajectories. Furthermore, a novel collision avoidance constraint is formulated for robot systems and obstacles of changeable shape. We validate our method through simulation comparison and real-world experiments, demonstrating its ability to guide the CT-QR system to target pose. Future work will involve cable tension control to improve motion accuracy, improving planning efficiency, and expand to multi-robot systems.



\ifCLASSOPTIONcaptionsoff
  \newpage
\fi



\bibliographystyle{IEEEtran}
\bibliography{references}

\end{document}